%% file: RTS_ICML_camera.tex
\documentclass{article}

\usepackage{times}
\usepackage{graphicx} 
\usepackage{subfigure} 

\usepackage{natbib}

\usepackage{algorithm}
\usepackage{algorithmic}

\usepackage{hyperref}

\usepackage[accepted] {sub/icml2016}

\usepackage{amsmath,amsfonts,amssymb,mathtools,bm}
\usepackage[margin=1in]{geometry}
\usepackage{algorithm,algorithmic}
\usepackage{float}
\usepackage{lipsum}

\icmltitlerunning{Partition Functions from Rao-Blackwellized Tempered Sampling }

\input{sub/definitions}

\begin{document} 

\twocolumn[
\icmltitle{Partition Functions from  \\ Rao-Blackwellized Tempered Sampling }

\renewcommand{\thefootnote}{\fnsymbol{footnote}}
\icmlauthor{David E. Carlson\footnotemark[1]$^{1,2}$}{david.edwin.carlson@gmail.com}
\icmlauthor{Patrick Stinson\footnotemark[1]$^{2}$}{patrickstinson@gmail.com}
\icmlauthor{Ari Pakman\footnotemark[1]$^{1,2}$}{ari@stat.columbia.edu}
\icmlauthor{Liam Paninski$^{1,2}$}{liam@stat.columbia.edu}
\icmladdress{$^1$ Department of Statistics\\
$^2$ Grossman Center for the Statistics of Mind\\
Columbia University, New York, NY, 10027}


\icmlkeywords{partition functions, MCMC , machine learning, ICML}

\vskip 0.3in
]
\renewcommand{\thefootnote}{\fnsymbol{footnote}}
\setcounter{footnote}{0}
\renewcommand{\thefootnote}{\arabic{footnote}}
\input{sub/abstract.tex}

\input{sub/intro.tex}

\input{sub/rtsIntro}

\input{sub/likelihood}

\input{sub/rtsDetails}

\input{sub/relatedWorkIntro}
\input{sub/mbar}

\input{sub/ti}

\input{sub/examplesIntro}

\input{sub/rts_rbm}

\input{sub/discussion}
\bibliography{RTS}
\bibliographystyle{sub/icml2016}

\newpage
\twocolumn[
\icmltitle{Partition Functions from  \\ Rao-Blackwellized Tempered Sampling: Supplemental Material }
\vskip 0.3in
]
\appendix
\input{sub/optimization}

\input{sub/rts_ti}
\input{sub/mbarProof}
\input{sub/hmc}
\input{sub/uniformExperiment}

\input{sub/transitionMatrix}

\end{document}

%% file: sub/definitions.tex
\newcommand{\eq}{\begin{equation*}}
\newcommand{\en}{\end{equation*}}
\newcommand{\eqa}{\begin{eqnarray*}}
\newcommand{\ena}{\end{eqnarray*}}
\newcommand{\eqn}{\begin{equation}}
\newcommand{\enn}{\end{equation}}
\newcommand{\be}{\begin{equation}}
\newcommand{\ee}{\end{equation}}
\newcommand{\eqan}{\begin{eqnarray}}
\newcommand{\enan}{\end{eqnarray}}

\newcommand{\hc}{ {\hat{c}} }

\newcommand{\x}{ {\bf x} }

\newcommand{\pmat}{\begin{pmatrix}}
\newcommand{\pman}{\end{pmatrix}}

\usepackage{amsthm}
\theoremstyle{plain}
\newtheorem*{theorem*}{Theorem}

%% file: sub/abstract.tex
\begin{abstract} 
Partition functions of probability distributions
are important quantities for model evaluation
and comparisons. We present a new method
to compute partition functions of complex and
multimodal distributions. Such distributions
are often sampled using simulated tempering,
which augments the target space with an
auxiliary inverse temperature variable. Our method exploits
the multinomial probability law of the
inverse temperatures, and provides estimates of the
partition function in terms of a simple quotient
of Rao-Blackwellized marginal inverse temperature
probability estimates, which are updated
while sampling. We show that the method
has interesting connections with several alternative
popular methods, and offers some significant advantages.
In particular, we empirically
find that the new method provides more accurate
estimates than Annealed Importance Sampling
when calculating partition functions of large
Restricted Boltzmann Machines (RBM); 
moreover, the method is sufficiently accurate to 
track training and validation log-likelihoods during learning of RBMs, at minimal computational cost.
\end{abstract}

%% file: sub/intro.tex
\section{Introduction}

The computation of partition functions (or equivalently, normalizing constants) and marginal likelihoods is an important 
problem in machine learning, statistics and statistical physics, 
and is necessary in tasks such as evaluating the test likelihood of complex generative models, 
calculating Bayes factors, or computing differences in free energies. 
There exists  a vast literature exploring methods to perform such computations, 
and the popularity and usefulness of different methods change across different communities and domain applications. 
Classic and recent reviews include \cite{gelman1998simulating, vyshemirsky2008bayesian, marin2009importance, friel2012estimating}.

In this paper we are interested in the  particularly challenging 
case of highly multimodal distributions, such as those common in machine learning applications~\cite{salakhutdinov2008quantitative}.
Our major novel insight is that simulated tempering, a popular approach for sampling  from such distributions, 
also provides an essentially cost-free way to estimate the partition function. 
Simulated tempering allows  sampling of multimodal distributions by augmenting the target space with a random inverse temperature variable and introducing a series of tempered distributions.  
The idea is that the fast MCMC mixing at low inverse temperatures allows the Markov chain to land in different modes of the low-temperature distribution of interest~\cite{marinari1992simulated, geyer1995annealing}.  

As it turns out, (ratios of) partition functions  have a simple expression in terms of ratios of the parameters of the multinomial probability law of the  inverse temperatures. 
These parameters can be estimated efficiently by averaging the conditional probabilities of the inverse temperatures along the Markov chain.
This simple method matches state-of-the-art performance with minimal computational and storage overhead.  
Since our estimator is based on Rao-Blackwellized marginal probability estimates of the inverse temperature variable, we denote it Rao-Blackwellized Tempered Sampling (RTS).

In Section~\ref{rts} we review the simulated tempering technique and introduce the new RTS estimation method.
In Section~\ref{sec3}, we compare RTS to Annealed Importance Sampling (AIS) and Reverse Annealed Importance Sampling (RAISE) \cite{ais,raise},
two popular methods in the machine learning community. 
We also show that RTS has  a close relationship with Multistate Bennett Acceptance Ratio (MBAR)~\cite{mbar,discriminance} and 
Thermodynamic Integration (TI)~\cite{gelman1998simulating}, two methods popular in the chemical physics and statistics communities, respectively.
In Section~\ref{examples}, we illustrate our method in a simple Gaussian example and in a Restricted Boltzmann Machine (RBM), where it 
is shown that RTS clearly dominates over the AIS/RAISE approach. 
We also show that RTS is sufficiently accurate to track training and validation log-likelihoods of RBMs during learning, at minimal computational cost.
We conclude in Section~\ref{discuss}.

%% file: sub/rtsIntro.tex
\vspace{-1mm}
\section{Partition Functions from Tempered Samples}
\vspace{-1mm}
\label{rts}
In this section, we start  by reviewing the tempered sampling approach and 
then introduce  our procedure to estimate partition functions.
We note that our approach is useful not only as a stand-alone method for estimating partition functions,
but is essentially free in any application using tempered sampling. 
In this sense it is similar to importance sampling approaches to computing partition functions (such as AIS).

\begin{figure*}[t!]
\centering
\includegraphics[height=3.55cm, width=0.24\textwidth]{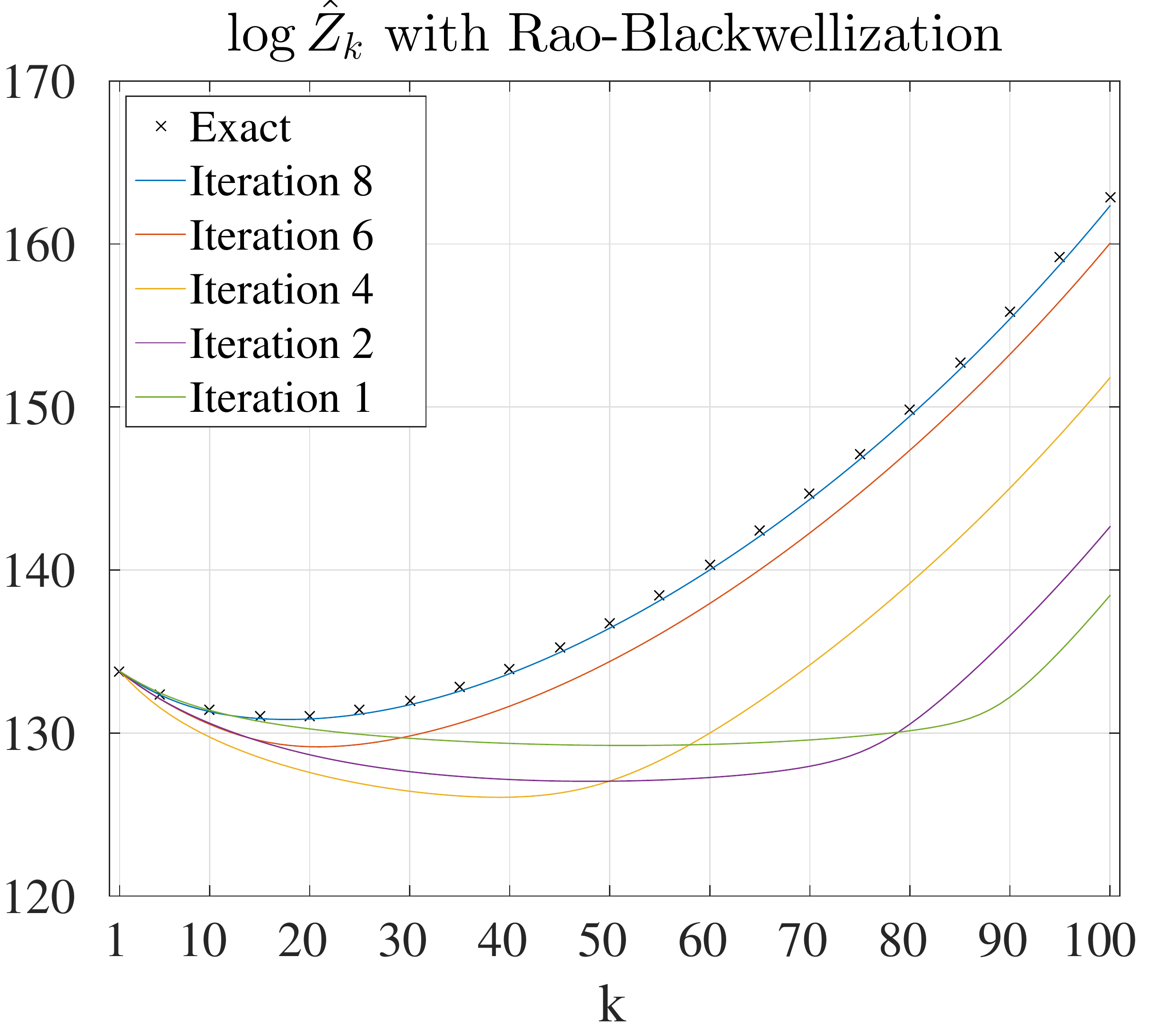}
\includegraphics[height=3.5cm,width=0.24\textwidth]{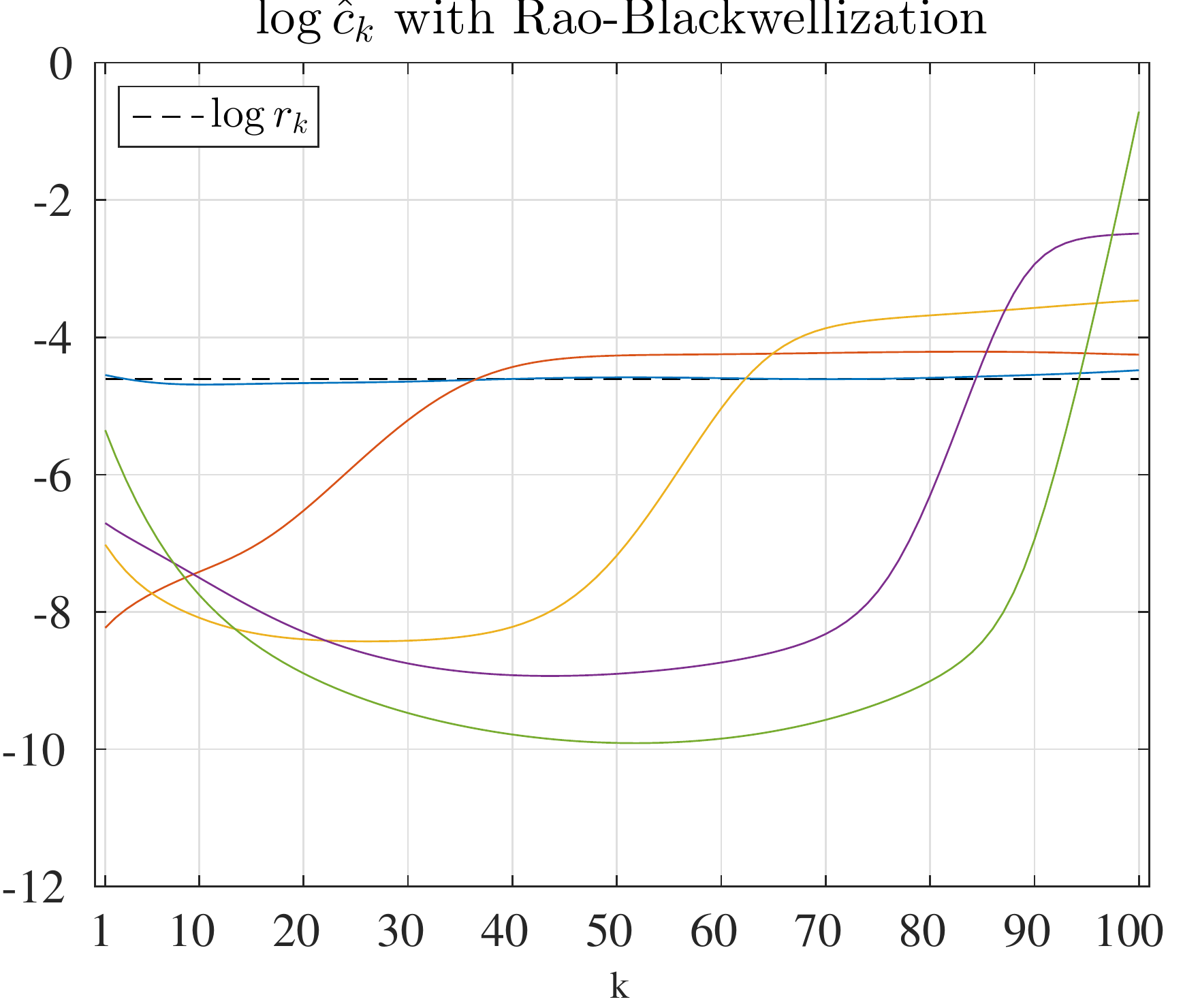}
\includegraphics[height=3.55cm, width=0.25\textwidth]{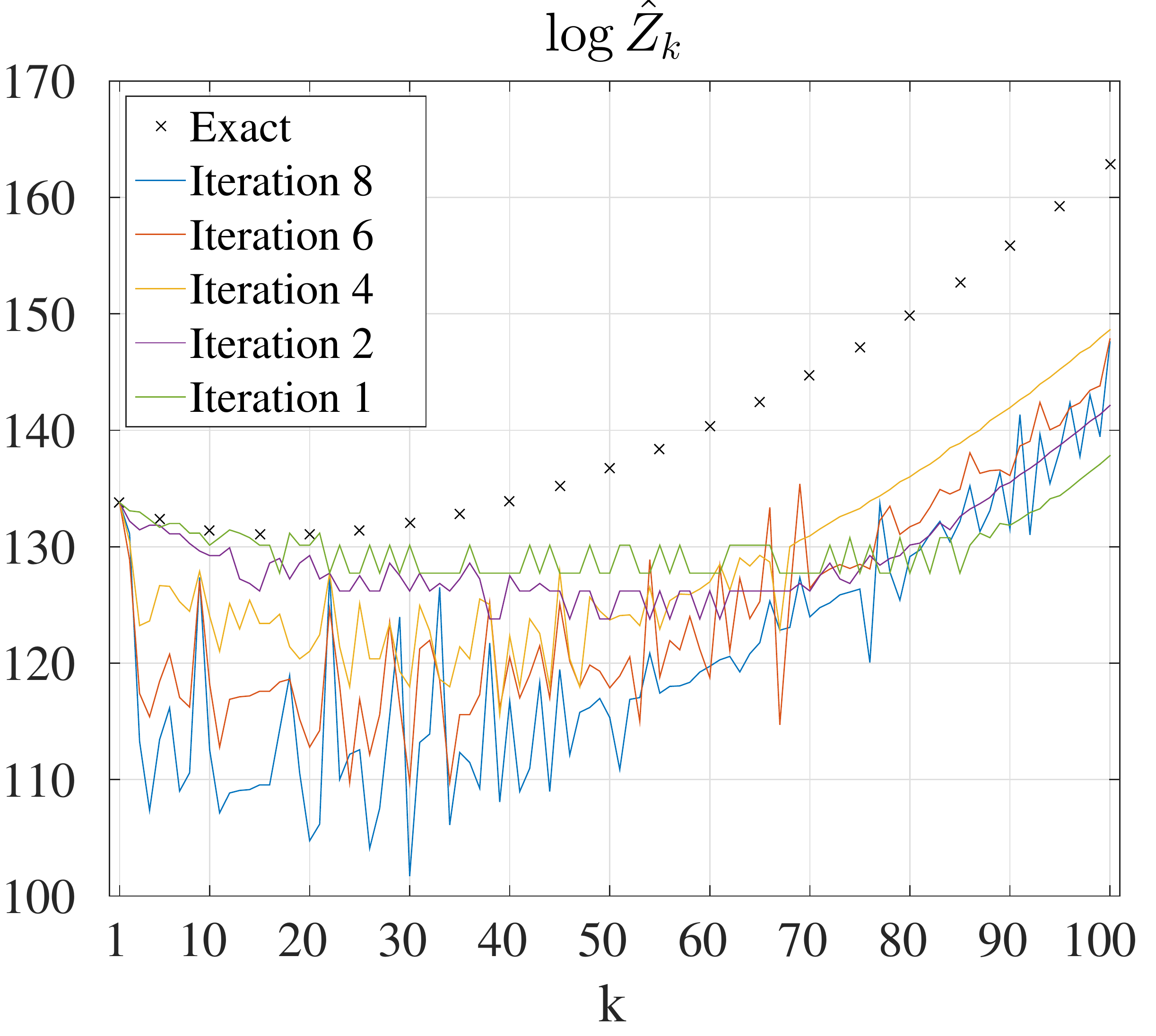}
\includegraphics[height=3.5cm,width=0.25\textwidth]{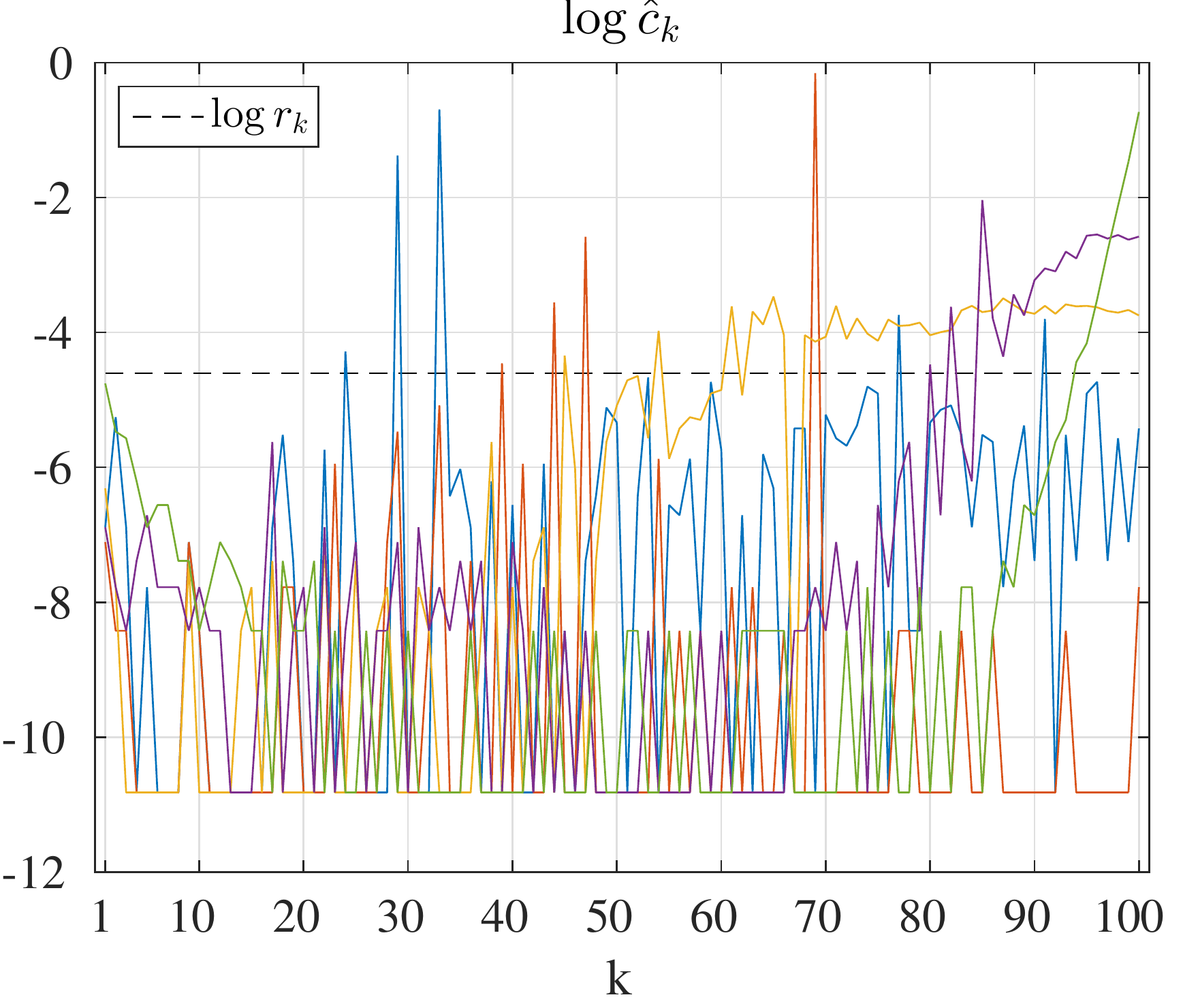}
\vspace{-2mm}
\caption{Comparison of $\log \hat{Z}_k$ and $\log \hat{c}_k$ estimates, in some of the first eight iterations of the initialization procedure described in Section~\ref{sec:burnin}, with and without Rao-Blackwellization, with $K=100$. 
The initial values were $\hat{Z}_k =1$ for all $k$, and the prior was uniform, $r_k = 1/K$.  
 The model is a RBM with 784 visible and 10 hidden units, trained on the MNIST dataset. Each iteration consists of 50 Gibbs sweeps, on each of 100 parallel chains. 
  Since in the  non-Rao-Blackwellized case, the updates are unstable and sometimes infinite, for demonstration purposes only, we define $\hc_k \propto 0.1 + \sum_{i=1}^N \delta_{k,k^{(i)} }$ and normalize. 
    Note that  in the Rao-Blackwellized case, the values of $\hc_k$ in the final iteration are very close to those of $r_k$, signaling that the $\hat{Z}_k$'s are 
 good enough for a last, long MCMC run to obtain the final~$\hat{Z}_k$ estimates.
  \vspace{-3mm}
  }
\label{iters}
\end{figure*}

\vspace{-2mm}
\subsection{Simulated Tempering}
\vspace{-1mm}
\label{sec:st}
Consider an unnormalized, possibly multimodal distribution proportional to $f(x)$, whose partition function we want to compute.
Our method is based on simulated tempering, a well known approach to sampling multimodal  distributions~\cite{marinari1992simulated,geyer1995annealing}.
Simulated tempering begins with a normalized and easy-to-sample distribution $p_1(x)$ 
and augments the target distribution with a set of discrete inverse temperatures $\{0=\beta_1<\beta_2<...<\beta_K=1\}$ to create a series of intermediate distributions between $f(x)$ and $p_1(x)$, given by 
\begin{flalign}
&\phantom{\hspace{1.5cm}}
p(x|\beta_k)=\textstyle \frac{f_k(x)} {Z_k}\,,\\
\text{where}&
\phantom{\hspace{1.5cm}}
\textstyle f_k(x)=f(x)^{\beta_k} p_1(x)^{1-\beta_k}\,,
\label{interpolating}
\\
\text{and}
\phantom{\hspace{.3cm}}&
\phantom{\hspace{1.5cm}}
\textstyle Z_k = \int f_k(x)  dx    \,.&
\end{flalign}
$Z_K$ is the normalizing constant that we want to compute.  
Note that we assume $Z_1=1$ and $p(x|\beta_1)=p_1(x)$.  
However, our method does not depend on this assumption.  When performing model comparison through likelihood ratios or Bayes factors, both distributions $f(x)$ and $p_1(x)$ can be unnormalized, and one is interested in the ratio of their partition functions.  For the sake of simplicity, we consider here only the interpolating family given in (\ref{interpolating}); other possibilities can be used for particular distributions, such as moment averaging~\cite{grosse2013annealing} or tempering by subsampling \cite{van2014tempering}.


When $\beta\in\{\beta_k\}_{k=1}^K$ is treated as a random variable, one can introduce a prior distribution $r(\beta_k)=r_k$, 
and define the joint distribution 
\eqan 
p(x,\beta_k) &=& p(x|\beta_k)r_k \,,
\\
&=&\textstyle \frac{f_k(x)r_k}{Z_k} \,.
\enan 
Unfortunately, $Z_k$ is unknown.  Instead, suppose we know approximate values $\hat{Z}_k$. Then we can define
\eqan 
q(x,\beta_k) &\propto& {f_k(x)r_k}/{\hat{Z}_k}\,,
\label{eqn:joint}
\enan 
which approximates $p(x,\beta_k)$.  
We note that the distribution $q$ depends explicitly on the parameters $\hat{Z}_k$.  A Gibbs sampler is run on this distribution by alternating between 
samples from  $x|\beta$ and $\beta|x$. The latter is given by 
\eqan 
q(\beta_k|x) = \frac{ f_k(x) r_k / \hat{Z}_k   } { \sum_{k'=1}^K  f_{k'}(x) r_{k'} /\hat{Z}_{k'}}  \label{pbx}   \,.
\enan 
Sampling as such enables the chain to traverse the inverse temperature ladder stochastically, escaping local modes under low $\beta$ and collecting samples from the target distribution $f(x)$ when $\beta=1$~\cite{marinari1992simulated}. When $K$ is large, few samples will have $\beta=1$.  Instead, an improved strategy to estimate expectations of functions over the target distribution is to Rao-Blackwellize, or importance sample, based on \eqref{pbx} to use all sample information \cite{geyer1995annealing}.
\vspace{-1mm}
\subsection{Estimating Partition Functions}
\vspace{-1mm}
Letting  $\hat{Z}_1 \equiv Z_1=1$, we first note that by  integrating  out $x$ in~(\ref{eqn:joint}) and normalizing, 
the marginal distribution over the~$\beta_k$'s is 
\eqan
\label{p_beta}
q(\beta_k)=\frac{r_k Z_k/\hat{Z}_k}{\sum_{k'=1}^K r_{k'} Z_{k'}/\hat{Z}_{k'}}  \,. \label{eq:qbeta}
\enan
Note that if $\hat{Z}_k$ is not close to $Z_k$ for all $k$, 
the marginal probability $q(\beta_k)$ will differ from the prior $r_k$, possibly by orders of magnitude for some $k$'s, and the $\beta_k$'s will not be efficiently sampled. 
One approach to compute approximate $\hat{Z}_k$ values  is the Wang-Landau algorithm~\cite{wang2001,atchade2010}.
We use an iterative strategy, discussed in Section \ref{sec:burnin}.

Given samples $\{x^{(i)}, \beta_{k^{(i)}} \}$ generated from $q(x,\beta_k)$, the marginal probabilities above
can simply be estimated by the normalized counts for each bin $\beta_k$, $\frac{1}{N}\sum_{i=1}^N \delta_{k,k^{(i)} }$.  
But a lower variance estimator  can be obtained by the Rao-Blackwellized form~\cite{robert2013monte}
\eqan 
\textstyle\hc_k = \frac{1}{N}\sum_{i=1}^N q(\beta_k|x^{(i)}) \,.
\label{ck}
\enan
The estimates in~(\ref{ck}) are unbiased estimators of~(\ref{p_beta}), since
\eqan 
\textstyle q(\beta_k)=\int q(\beta_k|x)q(x)dx\,.
\label{l_rb}
\enan 
Our main idea is that the exact partition function can be expressed by ratios of the marginal distribution in \eqref{p_beta},
\eqan 
Z_{k} = \hat{Z}_k \frac{r_1}{r_k}\frac{q(\beta_k)}{q(\beta_1)},     \qquad k = 2, \ldots, K \,.
\label{exactZ}
\enan 
Plugging our estimates~$\hc_k$ of $q(\beta_k)$ into~(\ref{exactZ}) immediately gives us the consistent estimator
\eqan 
 \hat{Z}_{k}^\text{RTS} = \hat{Z}_k \frac{r_1}{r_k}\frac{\hc_k}{\hc_1},     \qquad k = 2, \ldots, K \,.
 \label{solZ}
 \enan 
The resulting procedure is outlined in Algorithm~\ref{alg:example}.

%% file: sub/likelihood.tex
\vspace{-1mm}
\subsection{Rao-Blackwellized Likelihood Interpretation}
\vspace{-1mm}
\label{sec:opt_interp}
We can alternatively derive~(\ref{solZ}) by optimizing a Rao-Blackwellized form of the marginal likelihood. 
From (\ref{p_beta}), the log-likelihood 
of the $\{ \beta_{k^{(i)}} \}$ samples is 
\begin{align}
\log q( \{ \beta_{k^{(i)}} \}_{i=1}^N) &\textstyle = \sum_{i=1}^N   \log(Z_{k^{(i)}})  
\label{eqn:mle}
\\
\nonumber
 &\textstyle - N \log \left( \sum_{k=1}^K  r_{k} Z_{k}/\hat{Z}_{k}  \right) + const.
\end{align}
Because $\beta_{k^{(i)}}$ was sampled from $q(\beta|x^{(i)})$, we can reduce variance by Rao-Blackwellizing the first sum in  (\ref{eqn:mle}), resulting in
\begin{align} 
{L}_{RB}[\textbf{Z}] &\textstyle = \sum_{i=1}^N  \sum_{k=2}^K \log(Z_k)  q(\beta_k|x^{(i)}) 
\nonumber 
\\
&\textstyle - N \log \left( \sum_{k=1}^K  r_{k} Z_{k}/\hat{Z}_{k} \right) + const,
\nonumber 
\\
&\textstyle = 
N \sum_{k=2}^K \log(Z_k) \hc_k 
\label{raob}
\\
&\textstyle  - N \log \left( \sum_{k=1}^K  r_{k} Z_{k}/\hat{Z}_{k} \right)  + const \,.
\nonumber 
\end{align}
The normalizing constants are estimated by maximizing \eqref{raob} subject to a fixed $Z_1$, which is known.
Setting the derivatives of~(\ref{raob}) w.r.t. $Z_k$'s to zero gives a system of linear equations
\eqan
{\textstyle  \sum_{k'=2}^K}  \frac{r_{k'}}{\hat{Z}_{k'} } \left( \frac{ \delta_{k',k}}{\hc_k}   - 1 \right) Z_{k'} = r_1   \qquad k = 2, \ldots, K
\nonumber 
\label{linear_eqs}
\enan 
whose solution is~(\ref{solZ}).

%% file: sub/rtsDetails.tex
\begin{algorithm}[tb]
   \caption{Rao-Blackwellized Tempered Sampling 
   }
   \label{alg:example}
\begin{algorithmic}
   \STATE {\bfseries Input:} $\{\beta_k,r_k\}_{k=1,...,K},\, N$
   \STATE Initialize $\log\hat{Z}_k,\;k=2,...,K$
   \STATE Initialize $\beta\in\{\beta_1,...,\beta_K\}$
   \STATE Initialize $\hat{c}_k=0,\;k=1,...,K$
   \FOR{$i=1$ {\bfseries to} $N$}
   \STATE Transition in $x$ leaving $q(x|\beta)$ invariant.
   \STATE Sample $\beta|x\sim (\beta|x)$
   \STATE Update $\hat{c}_k\gets\hat{c}_k+\frac{1}{N}q(\beta_k|x)$
   \ENDFOR
   \STATE Update $\hat{Z}_k^{\text{RTS}}\gets \hat{Z}_k\frac{r_1 \hat{c}_k}{r_k\hat{c}_1} ,\;k=2,...,K$
\end{algorithmic}
\end{algorithm}

\vspace{-1mm}
\subsection{Initial Iterations}
\vspace{-1mm}
\label{sec:burnin}
As mentioned above, the chain with  initial $\hat{Z}_k$'s may mix slowly and provide a poor estimator (i.e. small $q(\beta_k)$'s are rarely sampled).  
Therefore, when the $\hat{Z}_k$'s are far from the $Z_k$'s (or equivalently, the $r_k$'s are far from the $\hat c_k$'s), the $\hat{Z}_k$'s estimates should be updated.  

Our estimator in \eqref{solZ} does not directly handle the case where $\hat{Z}_k$ is sequentially updated.  We note that the likelihood approach of \eqref{raob} is straightforwardly adapted to this case and is straightforwardly numerically optimized
(see Appendix  \ref{appdx:mixedzhat} for details). A simpler, less computationally intensive, and equally effective strategy is 
as follows: start with $\hat{Z}_k=1$ for all $k$ (or a better estimate, if known), and iterate between estimating $\hc_k$ with few MCMC samples
and updating $\hat{Z}_k$ with the estimated $\hat{Z}_{k}^\text{RTS}$ using (\ref{solZ}).
In our experiments using many parallel Markov chains, this procedure worked best when 
the updated Markov chains started from the previous last $x$'s, and fresh, uniformly random sampled $\beta_k$'s.

Once the $\hat{Z}_k$'s estimates are close enough to the $Z_k$'s to facilitate mixing, a long MCMC chain can be run to provide samples for the estimator.  
Because $\hc_k$ estimates $q(\beta_k)$,  and $q(\beta_k) \simeq r_k$ when 
$\hat{Z}_k \simeq {Z}_k$,  a simple stopping criterion for the initial iterations 
is to check the similarity between $\hc_k$ and $r_k$.  For example,  if we use a uniform
prior $r_k= 1/K$, a practical rule is to iterate the few-samples chains  until $\max_k |r_k - \hc_k| < 0.1/K$.

%
%

Figure~\ref{iters} shows the values taken by $\hat{Z}_k$ and $\hc_k$ in these initial iterations in a simple example. The figure also illustrates the importance
of using the Rao-Blackwellized form (\ref{ck}) for $\hc_k$, 
which dramatically reduces the noise in the  estimator $\frac{1}{N}\sum_{i=1}^N \delta_{k,k^{(i)} }$ for $q(\beta_k)$.

%
%

\subsection{Bias and Variance}
In Appendix \ref{appdx:biasandvar}, we show that the bias and variance of $\log\hat{Z}_k$ using Eqn. (\ref{solZ}) can be approximated by
\begin{flalign}
&\mathbb{E} \left[ \log \hat{Z}_k^\text{RTS} \right]-\log {Z}_k  \approx  {\textstyle\frac12 \left[  \frac{ \sigma^2_1}{\hc_1^2} -   \frac{ \sigma^2_k }{\hc_k^2}     \right]}\,,&\\
\label{bias}
\textstyle
\text{\hspace{-0mm}and}\phantom{\hspace{.5cm}}&\textstyle\text{Var}[\log\hat{Z}_k^\text{RTS}]\approx \frac{ \sigma^2_1}{\hc_1^2} +    \frac{ \sigma^2_k }{\hc_k^2}  -   \frac{ 2 \sigma_{1k} }{\hc_k \hc_1 }. &
\end{flalign}
where $\sigma^2_1=\text{Var}[\hat{c}_1]$, $\sigma^2_k=\text{Var}[\hat{c}_k]$, and $\sigma_{1k}=\text{Cov}[\hat{c}_1,\hat{c}_k]$.
This shows that the bias of $\log \hat{Z}_k$ has no definite sign.  
This is in contrast to many popular methods, such as AIS, which underestimates $\log Z_k$ \cite{ais}, and RAISE, which overestimates $\log Z_k$ \cite{raise}.

%% file: sub/relatedWorkIntro.tex
\section{Related Work}
\label{sec3}
In this section, we briefly review some popular estimators and explore their relationship to the proposed RTS estimator (\ref{solZ}).
All the estimators below use a family of tempered distributions, as appropriate for multimodal distributions. 
In some cases the temperatures are fixed parameters, while in others they are random variables. 
Note that RTS belongs to the latter group, and relies heavily on the random nature of the temperatures.

\subsection{Wang-Landau}

A well-known approach to obtain approximate values of the $Z_k$'s is the Wang-Landau algorithm~\cite{wang2001,atchade2010}.
The setting is similar to ours, but the  
algorithm constantly modifies the $\hat{Z}_k$'s  along the Markov chain as different  $\beta_k$'s are sampled.
The factors that change the $\hat{Z}_k$'s asymptotically  converge to 1. 
The resulting $\hat{Z}_k$ estimates are usually good enough to allow mixing in the $(x,\beta)$ space~\cite{salakhutdinov2010learning}, but are
too noisy for purposes such as likelihood estimation \cite{tan2015optimally}.

\subsection{AIS/RAISE}
\label{sec:ais}

Annealed Importance Sampling (AIS)~\cite{ais} is perhaps the most popular method in the machine learning literature to estimate $\log Z_K$. 
Here, one starts  from a sample $x_1$ from  $p_1(x)$, and samples a point $x_2$, using a transition function $K_2(x_2|x_1)$ that leaves $f_2(x)$ invariant.
The process is repeated until one has sampled $x_K$ using a transition function 
that leaves $f(x)$ invariant. The vector $(x_1,x_2,...,x_K)$ is interpreted as a sample from an importance distribution on an extended space,  
while the original distribution $p(x_K)$ can be similarly augmented into an extended space.  
The resulting importance weight can be computed in terms of quotients of the $f_k$'s, and provides an  
unbiased estimator for $Z_K/Z_1$, whose variance decreases linearly with $K$.
Note that the inverse temperatures in this approach are not random variables. 

The variance of the AIS estimator can be reduced 
by averaging over several runs, but the resulting value of $\log(\hat{Z}_K)$ has a negative bias due to Jensen's inequality. This in turn results in  a positive 
bias when estimating data log-likelihoods.

Recently, a related method, called Reverse Annealed Importance Sampling (RAISE)  was proposed to estimate the data log-likelihood in models with latent variables, 
giving  negatively biased estimates~\cite{raise, grosse2015sandwiching}. 
The method performs a similar sampling as AIS, but starts from a sample of the latent variables at $\beta_K=1$ and proceeds then to lower inverse temperatures. 
In certain cases, such as in the RBM examples we consider in Section~\ref{sec:pf_rbm}, one can obtain from these estimates of the data log-likelihood an estimate of the partition function, which will have a positive bias. 
The combination of the expectations of the AIS and RAISE estimators thus  `sandwiches' the exact value~\cite{raise, grosse2015sandwiching}.  

%% file: sub/mbar.tex
\subsection{BAR/MBAR}
Bennett's acceptance ratio (BAR)~\cite{bennett1976}, also called bridge sampling~\cite{meng1996simulating}, is based on the identity
\eqan 
\frac{Z_k}{Z_1} = \frac { \mathbb{E}_{p(x|\beta_1)} [ \alpha(x) f_k(x) ]  }  {  \mathbb{E}_{p(x|\beta_k)} [ \alpha(x) f_1(x) ] } \,,
\label{bridge}
\enan 
where $\alpha(x)$ is an arbitrary function such that $0 < \int f_1(x) f_k(x) \alpha(x)dx < \infty$, which can be chosen to minimize the asymptotic variance. BAR has been generalized to estimate partition functions when sampling from multiple distributions, a method termed the multistate BAR (MBAR)~\cite{mbar}.

Assuming that there are $n_k$ i.i.d. samples for each inverse temperature $\beta_k$ ($N$ samples $\{x_i\}_{i=1,\dots,N}$ in total), and $\Delta_x=\log f(x)-\log p_1(x)$, the MBAR 
partition function estimates can be obtained by maximizing the log-likelihood function~\cite{tan2012theory}:
\begin{align} 
L[\textbf{Z}]  &\hspace{-1mm}=\hspace{-1mm} \frac{1}{N}  {\textstyle \sum_{i=1}^{N} }\log\left( {\textstyle  \sum_{k=1}^K} \frac{n_k}{N}\exp(-\log Z_k+\beta_k\Delta_{x_i})\hspace{-.5mm}\right)\nonumber 
\\ &\hspace{-1mm}+\hspace{-1mm}{\textstyle  \sum_{r=1}^K} \frac{n_r}{N} \log Z_r.\label{eq:mbar}
\end{align} 
This method was recently rediscovered and shown to compare favorably against AIS/RAISE in~\cite{discriminance}.  MBAR has many different names in different literatures, e.g. unbinned weighted histogram analysis method (UWHAM) \cite{tan2012theory} and reverse logistic regression \cite{geyer1994estimating}.

Unlike RTS, MBAR does not use the form of $q(\beta)$ when estimating the partition function.  
As a price associated with this increased generality, MBAR requires the storage of all collected samples, and the estimator is calculated by finding the maximum of \eqref{eq:mbar}.  This likelihood function does not have an analytic solution, and Newton-Raphson was proposed to iteratively solve this problem, which requires $\mathcal{O}(NK^2 + K^3)$ per iteration.  While RTS is less general than MBAR, RTS has an analytic solution and only requires the storage of the  $\hat{c}_k$ statistics.  
We note that this objective function is very similar to the one  discussed in Appendix \ref{appdx:mixedzhat}
for combining different $\hat{Z}_k$'s.

Recent work has proposed a stochastic learning algorithm based on MBAR/UWHAM \cite{tan2016locally}, with  updates based on the sufficient statistics $\hat{c}_k$ given by 
\eqan
\log \hat{Z}_k^{(t+1)}=\log \hat{Z}_k^{(t)}+\gamma_t \left(\frac{\hat{c}_k}{r_k}-\frac{\hat{c}_1}{r_1}\right). \label{eq:smbar}
\enan
The  step size is recommended to be set to $\gamma_t=t^{-1}$.  
Note the similarity with our estimator  from \eqref{solZ} in log space,  with 
$\log\left(\frac{\hat{c}_k}{r_k}\right)-\log\left(\frac{\hat{c}_1}{r_1}\right)$ 
as the update. 
We empirically found that when the $\hat{Z}_k$'s  are far away from the truth, our update \eqref{solZ} dominates over~\eqref{eq:smbar}.  
Because the first order Taylor series approximation to our estimator is the same as the term in \eqref{eq:smbar}, when $\hat{c}_k\simeq r_k$ the updates will essentially only differ by the step size~$\gamma_t$. 

We also note that there is a particularly interesting relationship between the the cost function for MBAR and the cost function for RTS.  Note that $\mathbb{E}_q[\frac{n_k}{N}]$ is equal to $q(\beta_k)$ for tempered sampling.  If the values of $\frac{n_k}{N}$ in \eqref{eq:mbar} are replaced by their expectation, the maximizer of \eqref{eq:mbar} is equal to the RTS estimator given in \eqref{solZ}.  We detail this equivalency in Appendix \ref{mbar_equiv}.  Hence, the similarity of MBAR and RTS will depend on how far the empirical counts vary from their expectation. 
In our experiments, this form of extra information empirically helps to improve estimator accuracy.

%% file: sub/ti.tex
\vspace{-1mm}
\subsection{Thermodynamic Integration}
\vspace{-1mm}
Thermodynamic Integration (TI)~\cite{gelman1998simulating} is derived from basic calculus identities. Let us first assume that $\beta$ is a continuous variable in $[0,1]$. We again define $\Delta_x=\log f(x)-\log p_1(x)$, and $f_\beta(x)=f(x)^\beta p_1(x)^{1-\beta}$. We note that
\begin{align}
\frac{d}{d\beta}\log Z(\beta)&=\int \frac{1}{Z(\beta)}\frac{d}{d\beta}f_\beta(x)dx\nonumber\\
&=\mathbb{E}_{x|\beta}[\Delta_x],
\label{eqn:tiDeriv}
\end{align}
which yields
\eqan
\log\left(\frac{Z_K}{Z_1}\right)=\int_0^1 \mathbb{E}_{x|\beta}[\Delta_x]d\beta=\mathbb{E}_{p(x|\beta)p(\beta)}\left[ \frac{\Delta_x}{ p(\beta)}\right].\nonumber
\enan
This equation holds for any $p(\beta)$ that is positive over the range $[0,1]$, and provides an unbiased estimator for $\log Z_k$ if unbiased samples from $p(x|\beta)$ are available.  This is in contrast to AIS, which is unbiased on $Z_k$, and biased on $\log Z_k$.  Given samples $\{\x^{(i)},\beta^{(i)}\}_{i=1,...,N}$, the estimator for $\log Z_K$ is
\begin{align}
\widehat{\log Z_K}=\log Z_1+\frac{1}{N}{\textstyle\sum_{i=1}^N} \frac{\Delta_{x^(i)}}{p(\beta^{(i)})}.\nonumber
\end{align}
There are two distinct approaches for generating samples and performing this calculation in TI.  First, $\beta$ can be sampled from a prior $p(\beta)$, and samples are generated from $f_\beta(x)$ to estimate the gradient at the current point in $\beta$ space. A second approach is to use samples generated from simulated tempering, which can facilitate mixing. However, the effective marginal distribution $q(\beta)$ must be estimated in this case.

When $\beta$ consists of a discrete set of inverse temperatures, the integral can be approximated by the trapezoidal or Simpson's rule.  In essence, this uses the formulation in \eqref{eqn:tiDeriv}, and uses standard numerical integration techniques.  Recently, higher order moments were used to improve this integration, which can help in some cases \cite{friel2014}.  As noted by~\cite{calderhead2009estimating}, this discretization error can be expressed as a sum of KL-divergences between neighboring intermediate distributions.  If the KL-divergences are known, an optimal discretization strategy can be used.  However, this is unknown in general. 

While the point of this paper is not to improve the TI approach, we note that the Rao-Blackwellization technique we propose also applies to TI when using tempered samples.  This gives that the Monte Carlo approximation of the gradient \eqref{eqn:tiDeriv} is
\begin{align}
\left.\frac{d}{d\beta}\log Z(\beta)\right|_{\beta=\beta_k} \simeq{\textstyle  \sum_{i=1}^N} \frac{q(\beta_k|x_i) \Delta_{x_i}}{\sum_{j=1}^Nq(\beta_k|x_j)}.\label{eq:tirb}
\end{align}
This reduces the noise on the gradient estimates, and improves performance when the number of bins is relatively high compared to the number of collected samples.  We refer to this technique as TI-Rao-Blackwell (TI-RB).

TI-RB is further interesting in the context of RTS, because of a surprising relationship: in the continuous $\beta$ limit, RTS and TI-RB are \textit{equivalent} estimators.  However, when using discrete inverse temperatures, RTS does not suffer from the discretization error that TI and TI-RB do. 

We show the derivation of this relationship in Appendix \ref{ti_equiv}, but we give a quick description here.  First, let the inverse temperature $\beta$ take continuous values.  Replacing the index $k$ by $\beta$ in ~(\ref{solZ}), we note that the estimator for RTS can be written as:
\begin{align}
\textstyle  \log \left(\frac{\hat{Z}_K}{Z_1}\right)^{\hspace{-1mm}(RTS)}\hspace{-5mm} &=  \int_0^1 \frac{d}{d\beta}\left(\log \hat{c}_\beta - \log r_\beta + \log \hat{Z}_\beta\right)d\beta,
\nonumber 
  \\
  &=  
  \int_0^1 
   \frac{ \sum_i q(\beta|x_i) \Delta_{x_i} }{\sum_j q(\beta|x_j)}  d\beta\,.
   \label{TI}
\end{align}
Note that the integrand of \eqref{TI} is exactly identical to the TI-RB gradient estimate from the samples given in \eqref{eq:tirb}.  After integration, the estimators will be identical.

We stress that while the continuous formulation of RTS and TI-RB are equivalent in the continuous limit, in the discrete case RTS \emph{does not} suffer from discretization error.  
RTS is also limited to the case when samples are generated by the joint tempered distribution $q(x,\beta)$; however, because it does not suffer from discretization error, we empirically demonstate that RTS is much less sensitive to the number of temperatures compared to TI (see Section \ref{sec:numberTemps}).



Parallels between other methods and Thermodynamic Integration can be drawn as well. As noted in~\cite{neal2005estimating}, the log importance weight for AIS can be written as
\eqan
\textstyle  \log w = \sum_{k=2}^K (\beta_k-\beta_{k-1})\Delta_{x_k}
\enan
and thus can be thought of as a Riemann sum approximation to the numerical integral under a particular sampling approach.

%% file: sub/examplesIntro.tex
\section{Examples}

\label{examples}
In this section, we study the ability of RTS to estimate partition functions in a Gaussian mixture model and in 
Restricted Boltzmann Machines and compare to estimates from popular existing methods. We also study the dependence of several methods on the number $K$ of inverse temperatures, 
and show that RTS can provide estimates of train- and validation-set likelihoods during RBM training at minimal cost.  The MBAR estimates used for comparison in this section were calculated with the {\tt pymbar} package\footnote{Code available from \url{https://github.com/choderalab/pymbar}}.
\begin{figure}[t]
\centering
\includegraphics[width=0.33\textwidth]{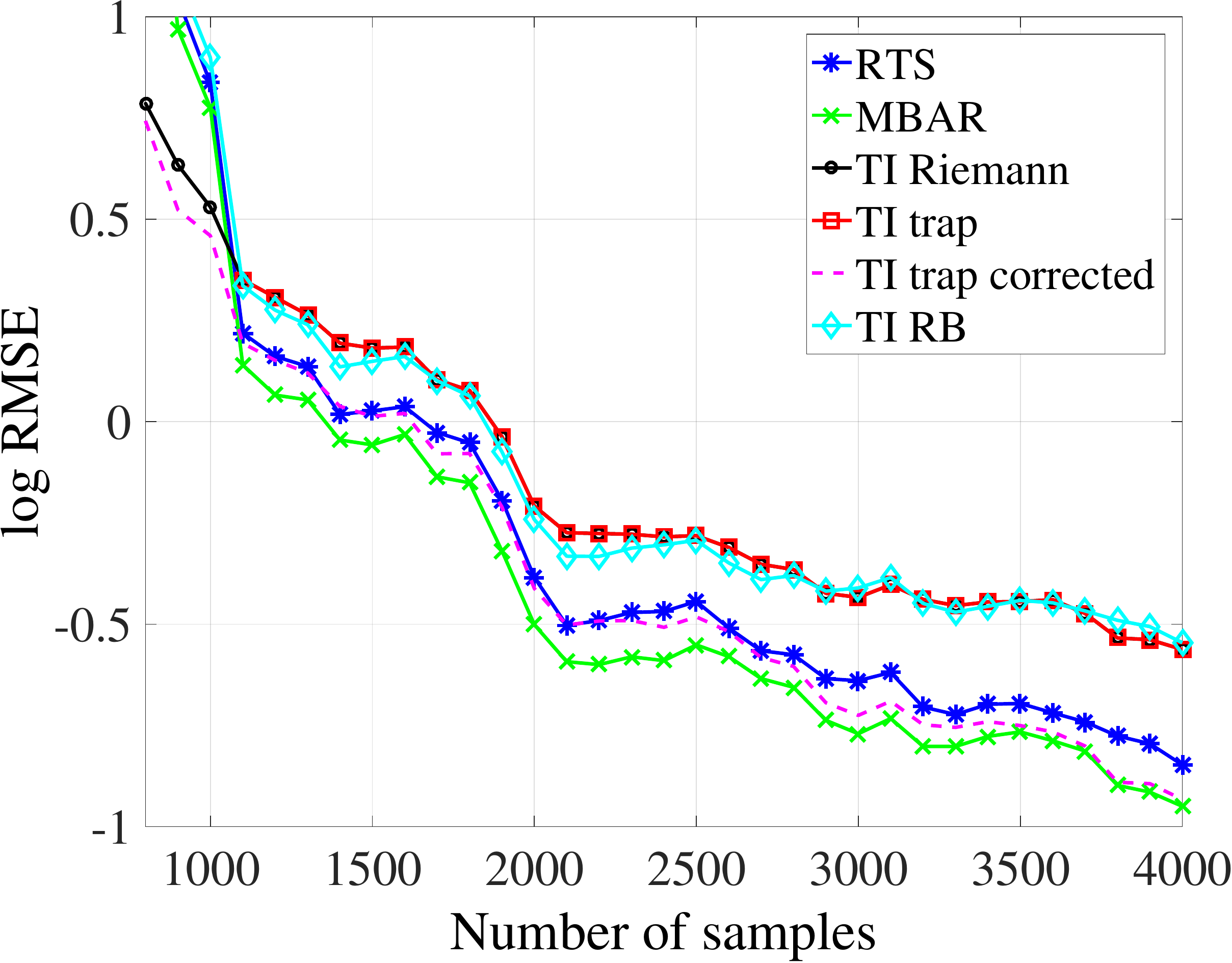}
\vspace{-2mm}
\caption{Comparison of $\log Z$ estimation performance on a toy Gaussian Mixture Model using an RMSE from 10 repeats. TI Riemann approximates the discrete integral as a right Riemann sum, TI trap uses the trapezoidal method, TI trap corrected uses a variance correction technique developed in~\cite{friel2014}, TI RB uses a Rao-Blackwellized version of TI discussed in Appendix~\ref{ti_equiv}. \vspace{-5mm}
}
\label{fig:gmm}
\end{figure}

\begin{figure*}[!t]
\includegraphics[width=.24\textwidth]{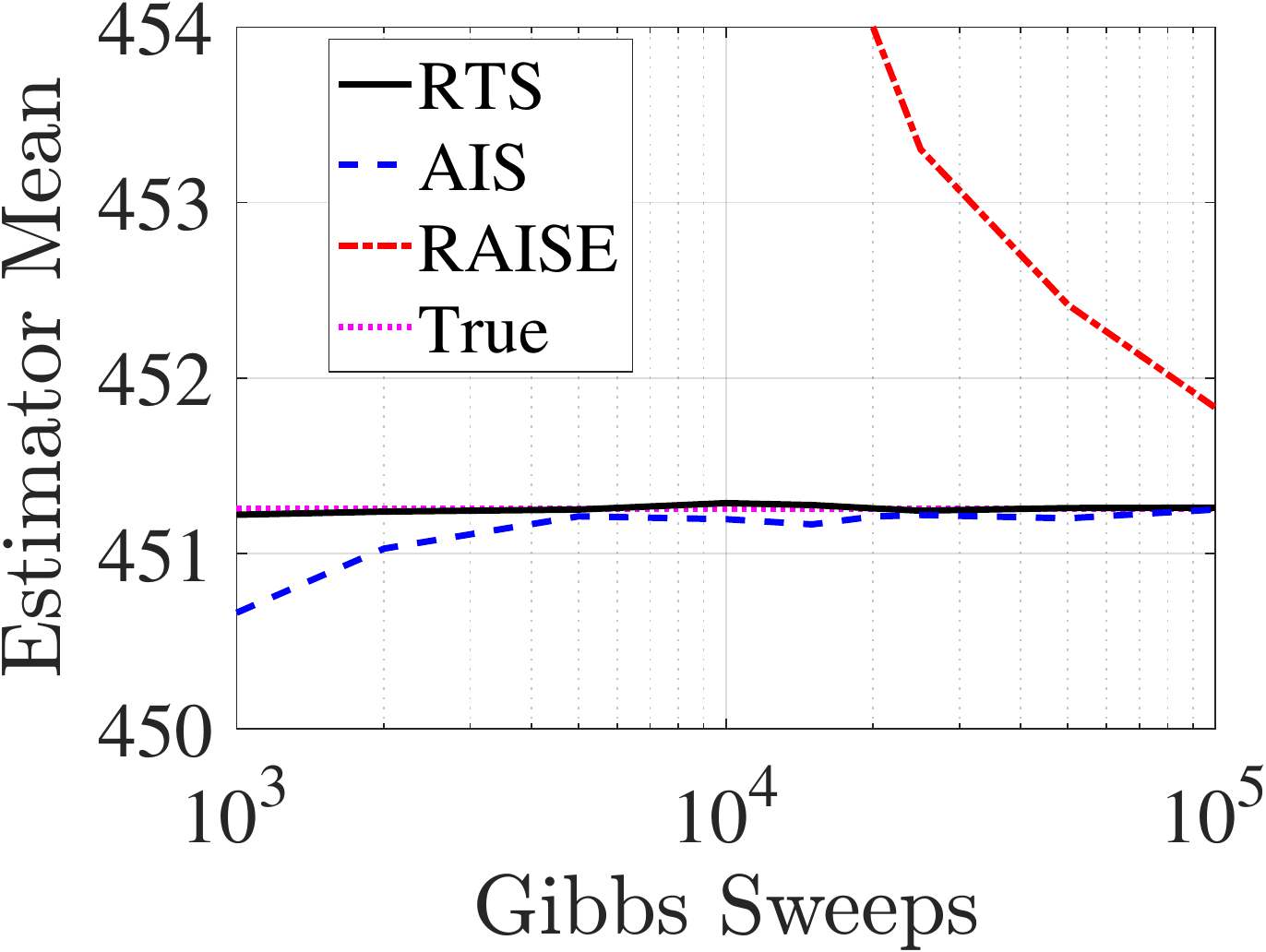}
\includegraphics[width=.24\textwidth]{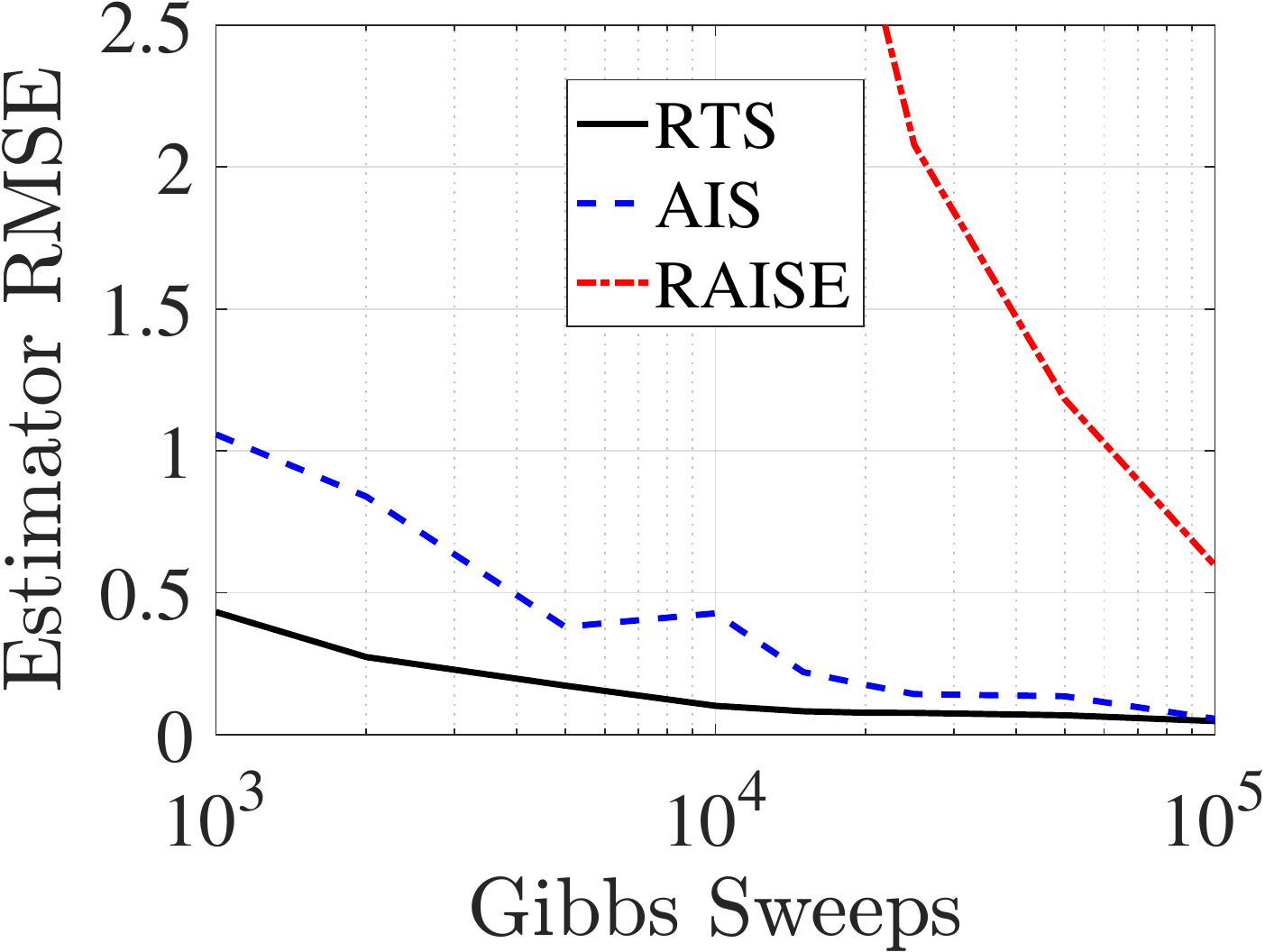}
\includegraphics[width=.24\textwidth]{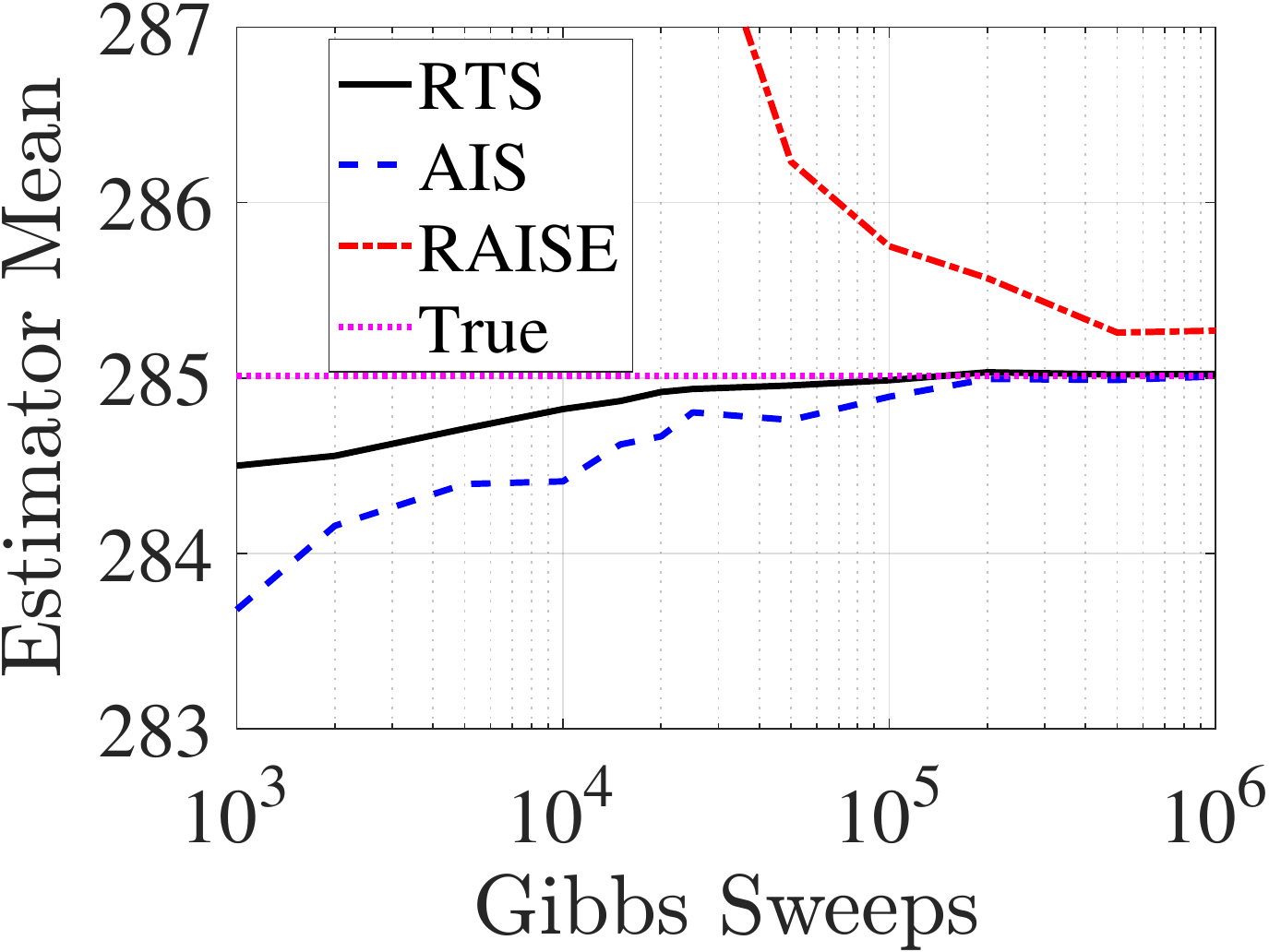}
\includegraphics[width=.24\textwidth]{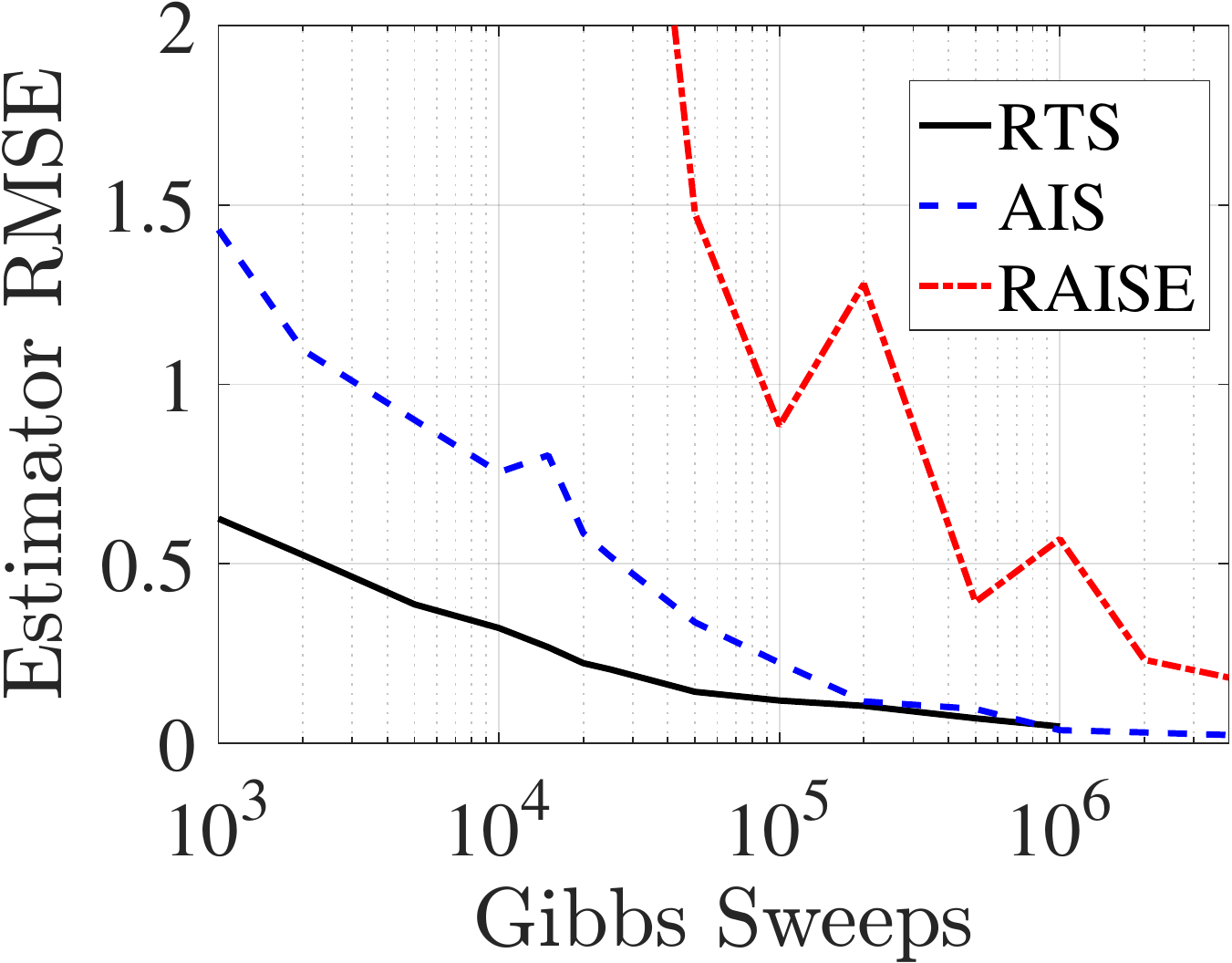}
\vspace{-1mm}
\caption{\label{fig:rbm} 
Mean and root mean squared error (RMSE) of competing estimators of $\log Z_K$ evaluated on RBMs with 784 visible units trained on the MNIST dataset. The numbers of hidden units were 500 (Left and Middle Left) and 100 (Middle Right and Right).
In both cases, the bias from RTS decreases quicker than that of AIS and RAISE, and the RMSE of AIS does not approach that of RTS at 1000 Gibbs sweeps until over an order of magnitude later.
Each method is run on 100 parallel Gibbs chains, but the Gibbs sweeps in the horizontal axis corresponds to each individual chain.
\vspace{-3mm}
}

\end{figure*}

\subsection{Gaussian Mixture Example and Comparisons}
Figure~\ref{fig:gmm} compares the performance of RTS to several methods, including MBAR and TI and its variants, 
in a  mixture of two 10-dimensional Gaussians (see Appendix \ref{sec:gaussian} for specific details).  
The sampling for all methods was performed  using a novel adaptive Hamiltonian Monte Carlo method for 
tempered distributions of continuous variables, introduced in Appendix \ref{sec:propadapt}. 
In this case the exact partition function can be numerically estimated to high precision. 
Note that all estimators give nearly identical performance; however, our method is the simplest to implement and use for tempered samples, with minimal memory and computation requirements.

%% file: sub/rts_rbm.tex
\subsection{Partition Functions of RBMs}
\label{sec:pf_rbm}

The Restricted Boltzmann Machine (RBM) is a bipartite Markov Random Field model popular in the machine learning community~\cite{smolensky1986information}.  
For the binary case, this is a generative model over  visible observations $v\in\{0,1\}^M$ and latent features $h\in\{0,1\}^J$ defined by  $\log f(v,h)= v^Tc + v^TWh+h^Tb$,  for parameters $c\in \mathbb{R}^M$, $b\in \mathbb{R}^J$, and $W\in\mathbb{R}^{M\times J}$.  
A fundamental performance measure of this model is the log-likelihood of a test set, which requires the estimation of the log partition function.  
Both AIS~\cite{salakhutdinov2008quantitative} and RAISE \cite{raise} were proposed to address this issue.    We will evaluate performance on the bias and the root mean squared error (RMSE) of the estimator. To estimate ``truth,'' we estimate the true mean as the average of estimates from AIS and RTS with $10^6$ samples from 100 parallel chains.  We note the variance of these estimates was very low ($\approx 0.006$).

Figure \ref{fig:rbm} shows a comparison of RTS versus AIS/RAISE on two RBMs trained on the binarized MNIST dataset ($M$=784, $N$=60000), 
with 500 and 100 hidden units. The former was taken from ~\cite{salakhutdinov2008quantitative},\footnote{Code and parameters available from: \url{http://www.cs.toronto.edu/~rsalakhu/rbm_ais.html} }
while the latter was trained with the method of~\cite{carlson2015preconditioned}.

In all the cases we used for $p_1$  a product of Bernoulli distributions over the $v$ variables which matches the marginal statistics of the training dataset, following~\cite{salakhutdinov2008quantitative}.
We run each method (RTS, AIS, RAISE) with 100 parallel Gibbs  chains.  
In RTS, the number of inverse temperatures was fixed at K=100, and we performed 10 initial iterations of 50 Gibbs sweeps each, following Section~\ref{sec:burnin}.
In AIS/RAISE, the number of inverse temperatures K was set to match in each case the total number of Gibbs sweeps in RTS, so the comparisons in Figure~\ref{fig:rbm} correspond to matched computational costs. 
We note that the performance of RAISE is similar to the plots shown in \cite{raise} for these parameters.  We also experimented with the case where $p_1$ was the uniform prior, and these results are included in Appendix \ref{sec:uniform}.


 \begin{figure}[!t]
 \centering
 \includegraphics[width=.4\textwidth]{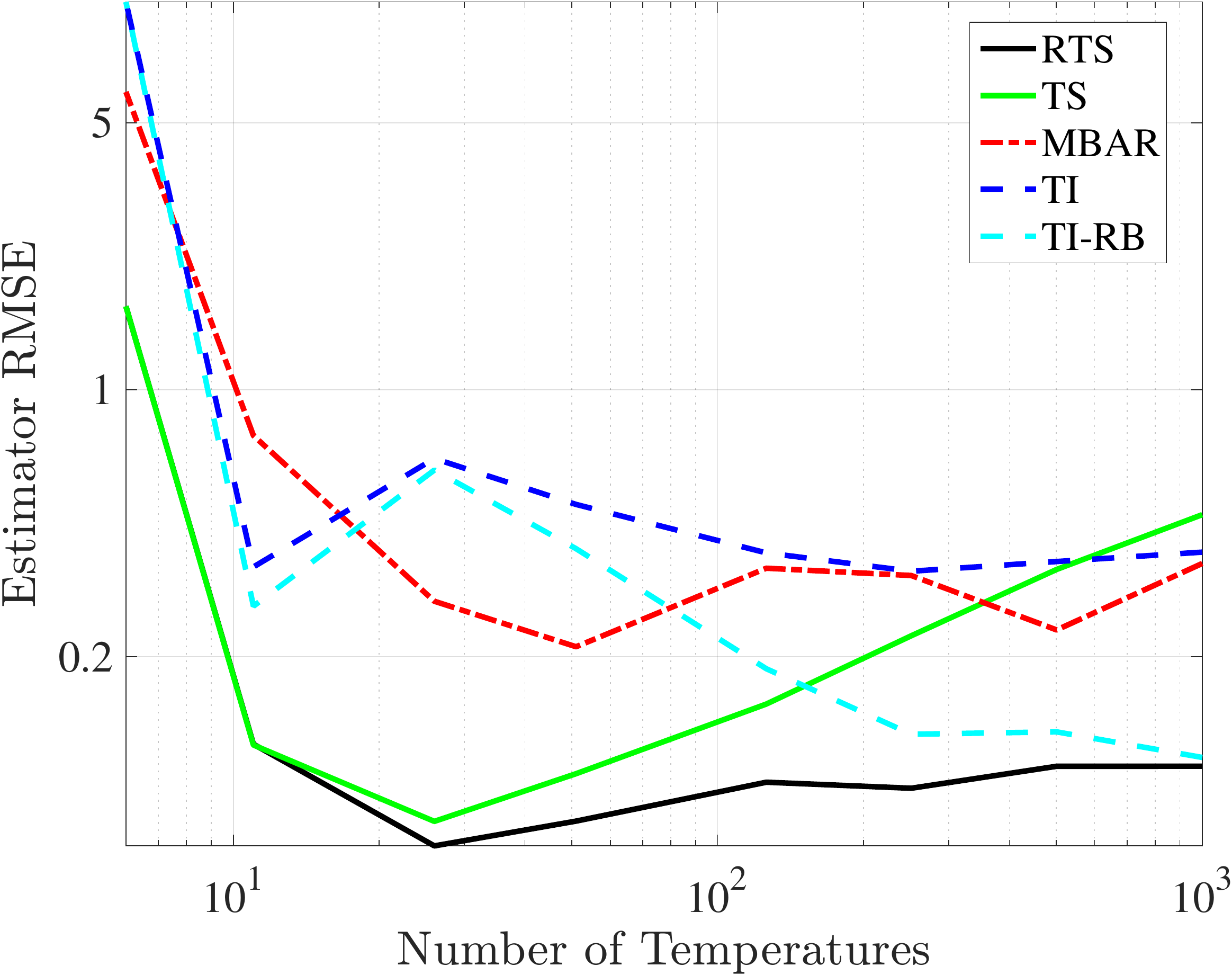}
 \caption{RMSE as a function of the number of inverse temperatures $K$ for various estimators. 
 The model is the same RBM with 500 hidden units studied in Figure~\ref{fig:rbm}. 
 Each point was obtained by averaging over 200 estimates (20 for MBAR due to computational costs) made from 10,000 bootstrapped samples from a long MCMC run of 3 million samples.
 \vspace{-5mm}}

 \label{numTemps}
 \end{figure}
\vspace{-1mm}
\subsection{Number of Temperatures}
\vspace{-1mm}
\label{sec:numberTemps}
 An advantage of the Rao-Blackwellization of temperature information is that there is no need to pick a precise number of inverse temperatures, as long as $K$ is big enough to allow for 
 good mixing of the Markov chain. As shown in Figure~\ref{numTemps}, RTS's performance is not greatly affected by adding more temperatures once there are enough temperatures to give good mixing. 
 
 Also note that as the number of temperatures increases RTS and the Rao-Blackwellized version of TI (TI-RB) become increasingly similar.  We show explicitly in Appendix \ref{ti_equiv} that they are equivalent in the infinite limit of the number of temperatures.
Due to computational costs, running MBAR on a large number of temperatures is computationally prohibitive. 
An issue when estimates are non-Rao-Blackwellized is that the estimates eventually become unstable as we do not have positive counts for each bin.  This is addressed heuristically in the non-Rao-Blackwellized version of RTS (TS) by adding a constant of $.1$ to each bin.  For TI, empty bins are imputed by linear interpolation.
%
%
\vspace{-1mm}
\subsection{Tracking Partition Functions While Training }
\vspace{-1mm}
\label{sec:tracking}
There are many approaches to training  RBMs, including  recent methods that do not require sampling~\cite{sohl2009minimum,im2014understanding,gabrie2015training}. 
However, most learning algorithms are based on Monte Carlo Integration with persistent Contrastive Divergence \cite{tieleman2009using}.
This includes proposals based on tempered sampling \cite{salakhutdinov2009learning,desjardins2010tempered}. 
Because RTS requires a relatively low number of samples and the parameters are slowly changing, we are able
to track the value of a train- and validation-set likelihoods during RBM training at minimal additional cost.
This allows us to avoid overfitting by early stopping of the training. 
We note that there are previous more involved efforts to track RBM  partition functions, which involve additional 
computational and implementation efforts~\cite{desjardins2011tracking}.

\begin{figure}
\centering
\includegraphics[width=.40\textwidth]{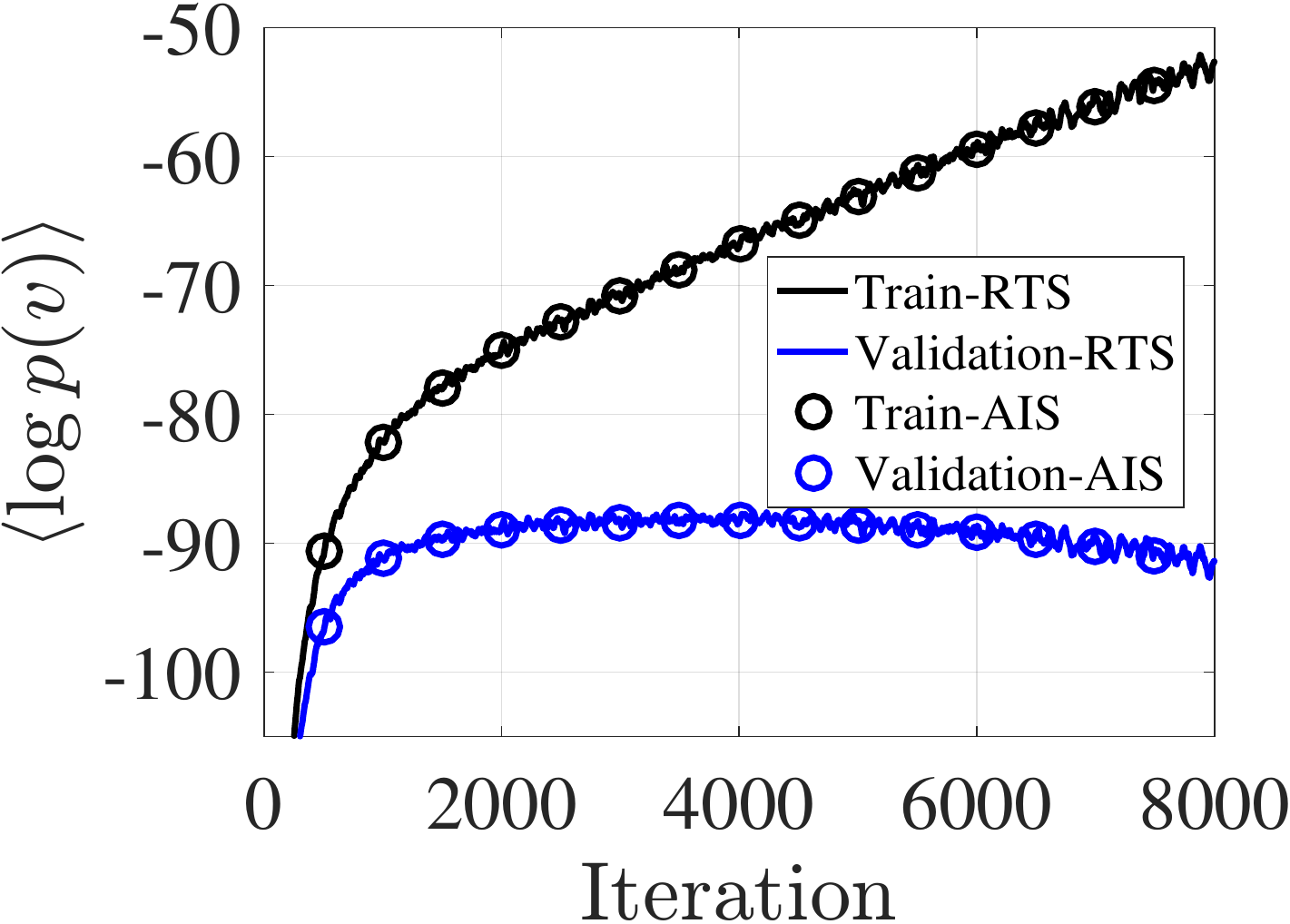}
\caption{\label{fig:rbmLearning} 
A demonstration of the ability to track with minimal cost  the mean train and validation log-likelihood during the training of a 
RBM on the {\it dna} 180-dimensional binary dataset, with 500 latent features.\vspace{-3mm}}
\end{figure}

This idea is illustrated in Figure \ref{fig:rbmLearning}, which shows 
estimates of the mean  of training and validation log-likelihoods on the {\it dna} dataset\footnote{Available from: \url{https://www.csie.ntu.edu.tw/~cjlin/libsvmtools/datasets/multiclass.html} },
with 180 observed binary features, trained on a RBM with  $500$ hidden units.

We first pretrain the RBM with CD-1 to get initial values for the RBM parameters. 
We then run initial RTS iterations with $ K=100$, as in Section~\ref{sec:burnin}, in order to get starting $\log \hat{Z}_k$  estimates. 

For the  main training effort we used the RMSspectral stochastic gradient method,
with stepsize of 1e-5 and parameter $\lambda=.99$ (see~\cite{carlson2015preconditioned} for details).
We considered a tempered space with $K=100$ and sampled 25 Gibbs sweeps on 2000 parallel chains between gradient updates.
The latter is a large number compared to older learning approaches \cite{salakhutdinov2008quantitative}, but is 
similar to 
that used both in 
\cite{carlson2015preconditioned} and \cite{grosse2015scaling} that provide state-of-the-art learning techniques.
We used a prior on the inverse temperatures $r_k\propto \exp(2\beta_k)$, which reduces variance on the gradient estimate by encouraging more of the samples to contribute to the gradient estimation.

With the samples collected after each 25 Gibbs sweeps, we can estimate the $\hc_k$'s to compute the running partition function. 
To smooth the noise from such a small number of samples, we consider partial updates of $\hat{Z}_K$ given by 
\eqan 
 \textstyle\hat{Z}_{K}^{(t+1)}  = \hat{Z}_K^{(t)}  \left( \frac{r_1}{r_K}   \frac{\hc_K^{(t)}}{\hc_1^{(t)}}   \right)^{\alpha}  
\enan 
with $\alpha=0.2$, and $t$  an index on the gradient update. Similar results were obtained with $.05<\alpha<.5$. This smoothing is also justified by the 
slowly  changing nature of the parameters. Figure~\ref{fig:rbmLearning} also shows the corresponding value from  AIS with 100 parallel samples and 10,000 inverse temperatures.  
Such AIS runs have been shown to give accurate estimates of the partition function for RBMs with even more hidden units~\cite{salakhutdinov2008quantitative}, 
but involve a major computational cost that our method avoids.  Using the settings from \cite{salakhutdinov2008quantitative} adds a cost of $10^6$ additional samples.

%


%
%
%

%% file: sub/discussion.tex
\vspace{-2mm}
\section{Discussion}
\vspace{-1mm}
\label{discuss}
In this paper, we have developed a new partition function estimation method that we called Rao-Blackwellized Tempered Sampling (RTS). 
Our experiments show RTS has equal or superior performance to  existing methods popular in the machine learning and physical chemistry communities,
while only requiring sufficient statistics collected during simulated tempering.

An important free parameter is the prior over inverse temperatures, $r_k$, and its optimal selection is a natural question. 
We explored several parametrized proposals for $r_k$, but in our experiments no alternative prior distribution consistently outperformed the uniform prior on estimator RMSE.  (In Section \ref{sec:tracking}, a non-uniform prior was used, but this was to reduce gradient estimate uncertainty at the expense of a less accurate $\log Z$ estimate.)
We also explored a continuous $\beta$ formulation, but the resulting estimates were less accurate. 
Additionally, we tried subtracting off estimates of the bias, but this did not improve the results.
Finally, we tried incorporating a variety of control variates, such as those in~\cite{dellaportas2012control}, but did not find them to reduce the variance of our estimates in the examples we considered.
Other control variates methods, such as those in~\cite{oates2015controlled}, could potentially be combined with RTS in continuous distributions.  We also briefly considered estimating $p(\beta_k)$ via the stationary distribution of a Markov process, which we discuss in Appendix \ref{sec:tm}.  This approach did not consistently yield performance improvements.
Future improvements could be obtained through improving the temperature path as in~\cite{grosse2013annealing,van2014tempering} or incorporating generalized ensembles \citep{frellsen2016bayesian}.

\section*{Acknowledgements}
We thank Ryan Adams for helpful conversations. Funding for this research was provided by DARPA N66001-15-C-4032 (SIMPLEX), a Google Faculty Research award, and ONR N00014-14-1-0243; in addition, this work was supported by the Intelligence Advanced Research Projects Activity (IARPA) via Department of Interior/ Interior Business Center (DoI/IBC) contract number D16PC00003.  The U.S. Government is authorized to reproduce and distribute reprints for Governmental purposes notwithstanding any copyright annotation thereon. Disclaimer: The views and conclusions contained herein are those of the authors and should not be interpreted as necessarily representing the official policies or endorsements, either expressed or implied, of IARPA, DoI/IBC, or the U.S. Government.

%% file: sub/optimization.tex
\section{Mixed $\hat{Z}$ Updates}
\label{appdx:mixedzhat}
We can generalize our Rao-Blackwellized maximum likelihood interpretation in Section~\ref{sec:opt_interp} to situations in which $\hat{Z}$ is not a fixed set of quantities for all samples. Under these conditions, we can no longer use the update in~(\ref{solZ}). However, we can easily find the Rao-Blackwellized log-likelihood, assuming independent~$\beta_k$ samples. Approximately independent samples can be obtained by sub-sampling with a rate determined by the autocorrelation of sampled $\beta$. We empirically found that  varying $\hat{Z}$ at late stages did not have a large effect on estimates.

Assume we have samples $\{x^{(i)},\beta^{(i)}\}$, with $\beta|x^{(i)}$ sampled using estimates $\hat{\textbf{Z}}^{(i)}=(\hat{Z}^{(i)}_k)_{k=1}^K$. Then our Rao-Blackwellized log-likelihood is the following 
\begin{align*}
L\left[\textbf{Z};\{\hat{\textbf{Z}}^{(i)}\}_{i=1}^N\right]&=\sum_{i=1}^N \sum_{k=2}^K \log Z_k q(\beta_k|x^{(i)};\hat{\textbf{Z}}^{(i)})\\&
-\sum_{i=1}^N \log\left( \sum_{k'=1}^K r_{k'}Z_{k'}/\hat{Z}_{k'}^{(i)}\right)\,,
\end{align*}
where
\begin{equation*}
q(\beta_k|x;\hat{\textbf{Z}}^{(i)})=\frac{f_k(x)r_k/\hat{Z}^{(i)}_k}{\sum_{k'=1}^K f_{k'}(x) r_{k'}/\hat{Z}^{(i)}_{k'}}\,.
\end{equation*}

Note that this expression is concave in $\log Z$ and can be solved efficiently using the generalized gradient descent methods of~\cite{carlson15,carlson16}.  The total computational time of this approach will scale $\mathcal{O}(K)$, whereas the Newton-Raphson method proposed in MBAR would scale $\mathcal{O}(K^3)$ per-iteration.  It is not clear how the number of iterations required in Newton-Raphson will scale, and could potentially have a worse dependence on $K$.

\section{Bias and Variance derivations}
\label{appdx:biasandvar}
A Taylor expansion of $\log \hat{Z}_k^\text{RTS}$, using (\ref{exactZ})-(\ref{solZ}) and $\log(1+x)  \simeq x-x^2/2$, gives
\eqan 
\log \hat{Z}^\text{RTS}_k  \approx \log {Z}_k +  \frac{\Delta c_k}{q_k}  - \frac{\Delta c_1}{q_1} -  \frac{(\Delta c_k)^2}{2 q_k^2}  + \frac{(\Delta c_1)^2}{2 q_1^2}   
\nonumber 
\enan 
where $q_k= q(\beta_k)$ and $\Delta c_k = \hat{c}_k - q_k$. 
Taking expectations, and replacing $q_k$ by its estimate $\hc_k$, gives
\eqan
\mathbb{E} \left[ \log \hat{Z}^\text{RTS}_k \right]-\log {Z}_k  \approx  \frac12 \left[  \frac{ \sigma^2_1}{\hc_1^2} -   \frac{ \sigma^2_k }{\hc_k^2}     \right]\,,
\enan
and 
\eqan
\text{Var}[\log\hat{Z}_k^\text{RTS}]\approx \frac{ \sigma^2_1}{\hc_1^2} +    \frac{ \sigma^2_k }{\hc_k^2}  -   \frac{ 2 \sigma_{1k} }{\hc_k \hc_1 } 
\enan 
where $\sigma^2_1=\text{Var}[\hat{c}_1]$, $\sigma^2_k=\text{Var}[\hat{c}_k]$, and $\sigma_{1k}=\text{Cov}[\hat{c}_1,\hat{c}_k]$.

From the CLT, the asymptotic variance of $\hat{c}_k$ is 
\eqan 
Var(\hat{c}_k) = \frac{ Var_q( q(\beta_k |x ) ) a_k}{N} \,,
\enan 
where the factor 
\eqan 
a_k = 1 + 2\sum_{i=1}^{\infty} \textrm{corr} \left[ q(\beta_k |x^{(0)} ), q(\beta_k |x^{(i)} ) \right]
\enan 
takes into account the autocorrelation of the Markov chain. But estimates of this sum from the MCMC samples are generally 
too noisy to be useful. 
Alternatively, $Var[\hat{c}_k]$ could simply be estimated from $\hc_k$ estimates on many parallel MCMC chains. 

%
%

%% file: sub/rts_ti.tex
\section{RTS and TI-RB Continuous $\beta$ Equivalence}
\label{ti_equiv}
We want to show the relationship mentioned in \eqref{TI}, which we repeat here:
\begin{align}
  \log \left(\frac{\hat{Z}_K}{Z_1}\right)^{\hspace{-1mm}(RTS)}\hspace{-5mm} &=  \int_0^1 \frac{d}{d\beta}\left(\log \hat{c}_\beta - \log r_\beta + \log \hat{Z}_\beta\right)d\beta,
\nonumber 
  \\
  &=  
  \int_0^1 
   \frac{ \sum_i q(\beta|x_i) \Delta_{x_i} }{\sum_j q(\beta|x_j)}  d\beta\,.\nonumber
\end{align}

Note that we can write the statistics $c_k$ as
\begin{align}
 c_k& = \sum_{i=1}^N q(\beta_k|x_i)\nonumber\\
 &=\sum_{i=1}^N\frac{\exp\left(\beta_k \Delta_{x_i}+\log r_k -\log \hat{Z}_k\right)}{\sum_{k'=0}^K\exp\left(\beta_{k'}\Delta_{x_i}+\log r_{k'} -\log \hat{Z}_{k'}\right)}\nonumber
 \end{align}
 The continuous version of this replaces the index $k$ by $\beta$, and
\begin{align}
 c_\beta &= \sum_{i=1}^N q(\beta|x_i)\nonumber
 \\&=\sum_{i=1}^N\frac{\exp\left(\beta \Delta_{x_i}+\log r_\beta -\log \hat{Z}_\beta\right)}{\int_{0}^1\exp\left(\alpha\Delta_{x_i}+\log r_{\alpha} -\log \hat{Z}_{\alpha}\right)d\alpha}\nonumber
 \end{align}
 The continuous form of the RTS estimator can be written as an integral:
\begin{align}
\log \frac{Z_K}{Z_1}&=\left.\left(\log c_\beta - \log r_\beta + \log \hat{Z}_\beta\right)\right|_{\beta=1}\nonumber\\
&-\left.\left(\log c_\beta - \log r_\beta + \log \hat{Z}_\beta\right)\right|_{\beta=0}\nonumber\\
&=\int_0^1 \frac{d}{d\beta}\left(\log c_\beta - \log r_\beta + \log \hat{Z}_\beta\right)d\beta\label{eq:tiInt}
\end{align}
We first analyze the derivative of $c_\beta$, which is 
\begin{align}
  &\frac{d}{d\beta}\log c_\beta\nonumber&\\
   =&\frac{d}{d\beta} \log\sum_{i=1}^N\frac{\exp\left(\beta \Delta_{x_i}+\log r_k -\log \hat{Z}_k\right)}{\int_0^1\exp\left(\alpha\Delta_{x_i}+\log r_{\alpha} -\log z_{\alpha}\right)d\alpha}&\nonumber\\
 =&\frac{1}{\sum_{i=1}^N\frac{\exp\left(\beta \Delta_{x_i}+\log r_\beta -\log \hat{Z}_\beta\right)}{\int_0^1\exp\left(\alpha\Delta_{x_i}+\log r_{\alpha} -\log \hat{Z}_{\alpha}\right)d\alpha}}\nonumber&\\
\times& \sum_{i=1}^N\frac{\exp\left(\beta \Delta_{x_i}+\log \frac{r_\beta}{\hat{Z}_\beta}\right)\frac{d}{d\beta}\left(\beta \Delta_{x_i}+\log \frac{r_\beta}{\hat{Z}_\beta}\right)}{\int_0^1\exp\left(\alpha\Delta_{x_i}+\log r_{\alpha} -\log \hat{Z}_{\alpha}\right)d\alpha}&\nonumber\\  =&\sum_i \frac{q(\beta|x_i)\frac{d}{d\beta}\left(\beta \Delta_{x_i}+\log r_\beta -\log \hat{Z}_\beta\right)}{\sum_j q(\beta|x_j)} \nonumber\\
  =&\left[\sum_i \frac{q(\beta|x_i)}{\sum_j q(\beta|x_j)} \Delta_{x_i}\right] +\frac{d}{d\beta}(\log r_\beta - \log \hat{Z}_\beta)\label{eq:tiFinal}
\end{align}
The last line follows since $\sum_{i=1}^N \frac{q(\beta|x_i)}{\sum_j q(\beta|x_j)}=1$.  The $\frac{d}{d\beta}(\log r_\beta - \log \hat{Z}_\beta)$ term in \eqref{eq:tiInt} and \eqref{eq:tiFinal} simply cancel. 

%% file: sub/mbarProof.tex
\section{Similarity of RTS and MBAR}
\label{mbar_equiv}
In this section, we elaborate on the similarity of the likelihood of MBAR and RTS.
To prove this, we first restate the likelihood of MBAR given in \eqref{eq:mbar}:
\begin{align}
L[\bf{Z}]&=\frac{1}{N}\sum_{i=1}^N \log \left( \sum_{k=1}^K \frac{n_k}{N}\exp(-\log Z_k + \beta_k \Delta_{x_i})\right)\nonumber\\&+\sum_{k=1}^N \frac{n_k}{N}\log Z_k\nonumber
\end{align}
The partial derivative of this likelihood with respect to $\log Z_k$ is given by:
\begin{align}
\tiny\frac{\partial L[\bf{Z}]}{\partial\log Z_k}&=\frac{n_k}{N}\label{eq:mbarderiv}\\
&-\frac{1}{N}\sum_{i=1}^N \frac{\frac{n_k}{N}\exp(-\log Z_k + \beta_k \Delta_{x_i})}{{\displaystyle \sum_{j=1}^K \frac{n_j}{N}}\exp(-\log Z_j + \beta_j \Delta_{x_i})}\nonumber
\end{align}
Replacing $\frac{n_k}{N}$ with its expectation for all $k$ gives
\begin{align}
\tiny\frac{\partial L[\bf{Z}]}{\partial\log Z_k}&=q(\beta_k)\label{eq:Embarderiv}\\
&-\frac{1}{N}\sum_{i=1}^N \frac{q(\beta_k)\exp(-\log Z_k + \beta_k \Delta_{x_i})}{\displaystyle \sum_{j=1}^K q(\beta_j)   \exp(-\log Z_j + \beta_j \Delta_{x_i})}\nonumber
\end{align}
Noting that $q(\beta_k)\propto Z_k/\hat{Z}_kr_k$, we have
\begin{align}
\tiny\frac{\partial L[\bf{Z}]}{\partial\log Z_k}&=q(\beta_k)\nonumber\\
&-\frac{1}{N} \sum_{i=1}^N \frac{\frac{Z_k}{\hat{Z}_k}r_k\exp(-\log Z_k+\beta_k\Delta_{x_i})}{\sum_{j=1}^K \frac{Z_j}{\hat{Z}_j}r_j\exp(-\log Z_j+\beta_j\Delta_{x_i})}\nonumber,\\
&=q(\beta_k)\nonumber\\
&-\frac{1}{N}\sum_{i=1}^N \frac{\exp(-\log\hat{Z}_k+\beta_k\Delta_{x_i})}{\sum_{j=1}^K\exp(-\log\hat{Z}_j+\beta_j\Delta_{x_i})}\nonumber,\\
&=q(\beta_k)-\frac{1}{N}\sum_{i=1}^N q(\beta_k|x_i)\nonumber,\\
&=q(\beta_k)-\hat{c}_k \label{eq:lastmbar}\,.
\end{align}
Setting the partial derivative to 0 and substituting the definition of $q(\beta)$ into \eqref{eq:lastmbar} gives a solution of
\begin{equation}
\frac{Z_k/\hat{Z}_kr_k}{\sum_{j=1}^K Z_j/\hat{Z}_jr_j}=\hat{c}_k,
\end{equation}
which is identical to the RTS update in \eqref{solZ}.

While RTS and MBAR give similar estimators, their intended use is a bit different.  The MBAR estimator can be used whenever we have samples generated from a distribution at different temperatures, including both physical experiments where the temperature is an input and a tempered MCMC scheme.
The RTS estimator requires a tempered MCMC approach, but in exchange has trivial optimization costs and improved empirical performance.

%% file: sub/hmc.tex
\begin{figure*}[t]
\centering
	\includegraphics[width=0.3\textwidth]{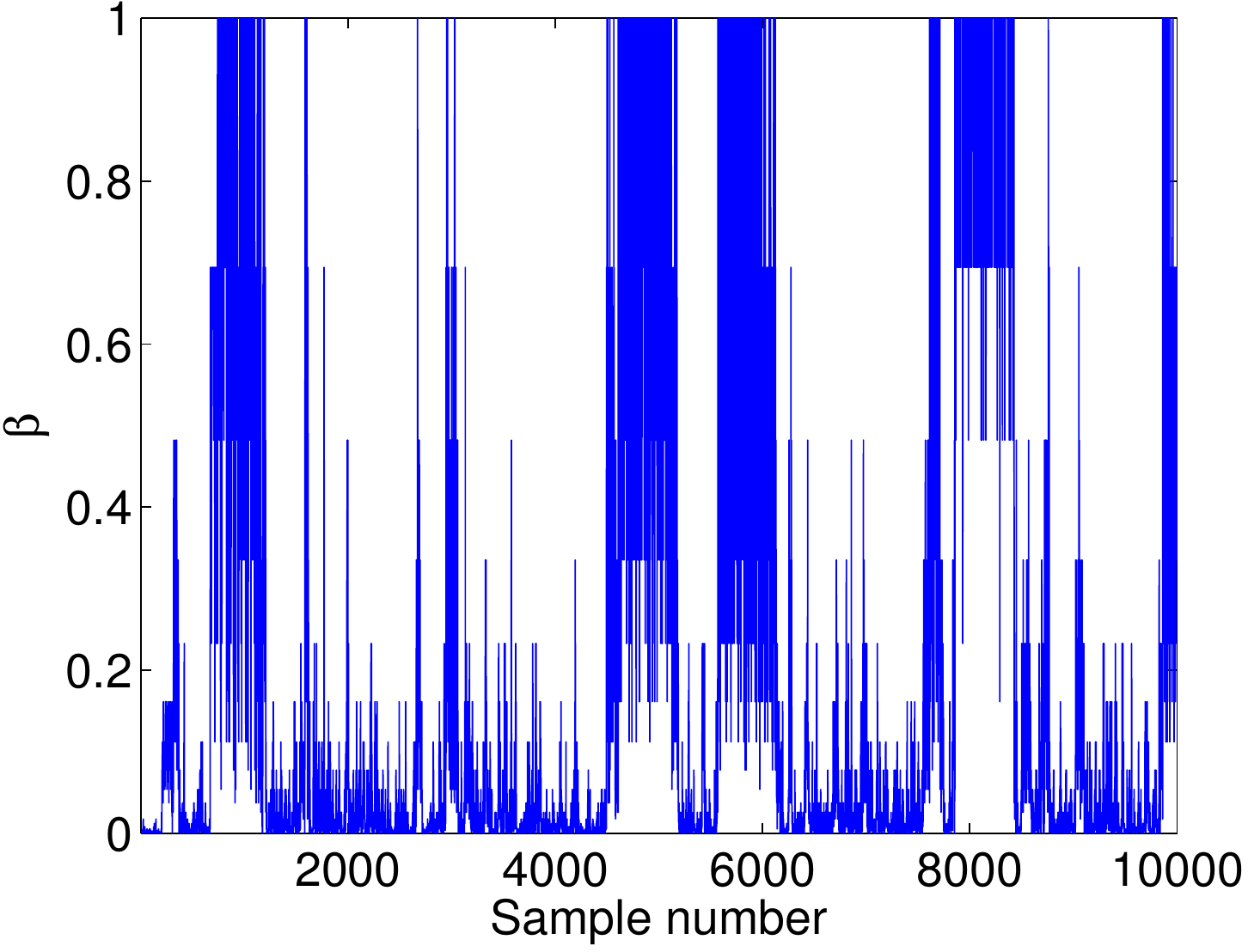}
	\includegraphics[width=0.3\textwidth]{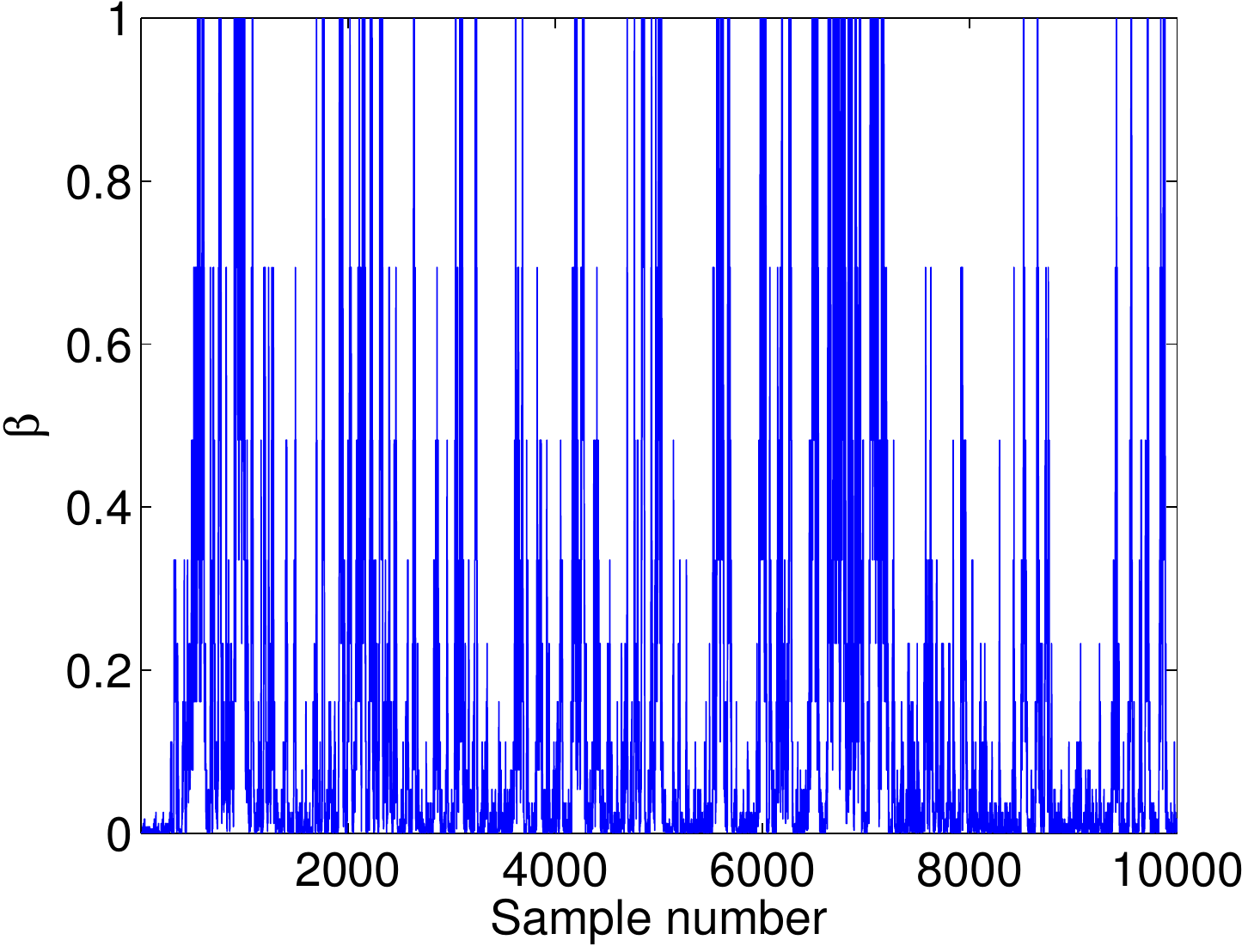}
	\includegraphics[width=0.3\textwidth]{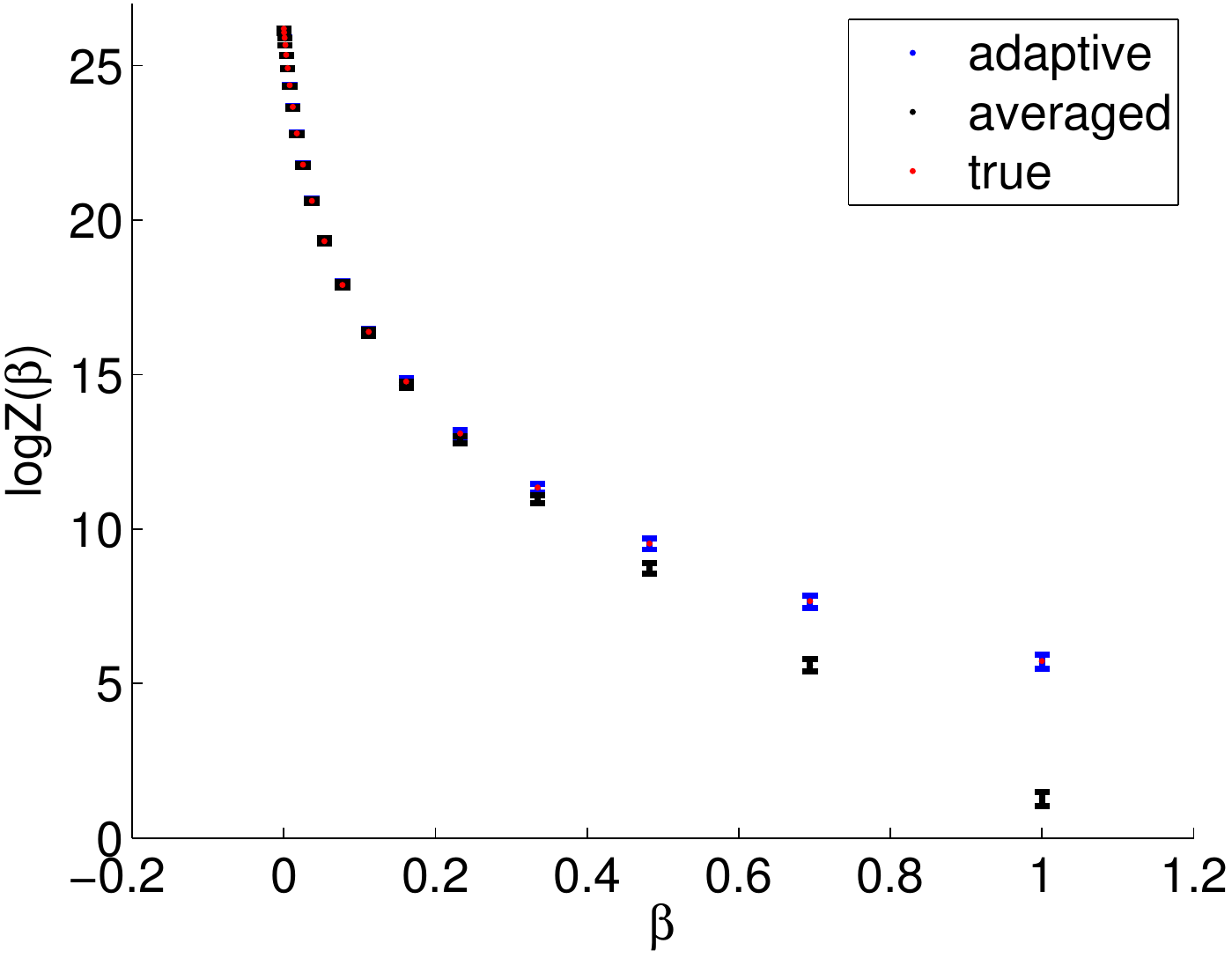}
	\caption{(Left) Mixing in $\beta$ under the fixed step size. (Middle) Mixing in $\beta$ under the adaptive scheme. (Right) Partition function estimates under the fixed step size and adaptive scheme after 10000 samples. Mixing in $\beta$ using a fixed step size is visibly slower than mixing using the adaptive step size, which is reflected by the error in the partition function estimate.}
\end{figure*}

\section{Adaptive HMC for tempering}
\label{sec:propadapt}
Here we consider sampling from a continuous distribution using Hamiltonian Monte Carlo (HMC)~\cite{neal_hmc}. Briefly, HMC simulates Hamiltonian dynamics as a proposal distribution for Metropolis-Hastings (MH) sampling. In general, one cannot simulate exact Hamiltonian dynamics, so usually one uses the leapfrog algorithm, a first order discrete integration scheme which maintains the time-reversibility and volume preservation properties of Hamiltonian dynamics.

\cite{ast} found using different step sizes improved sampling various multimodal distributions using random walk Metropolis proposal distributions. However, under their scheme, besides step sizes being monotonically decreasing in $\beta$, it is unclear how to set these step sizes. Additionally, in target distributions that are high-dimensional or have highly correlated variables, random walk Metropolis will work badly.

For most distributions of interest, as $\beta$ decreases, $p(x|\beta)$ becomes flatter; thus, for HMC, we can expect the MH acceptance probability to decrease as a function of $\beta$, enabling us to take larger jumps in the target distribution when the temperature is high. 
As the stepsize $\epsilon$ of the leapfrog integrator gets smaller, the linear approximation of the solution to the continuous differential equations becomes more accurate, and the MH acceptance probability increases (for an infinitely small stepsize, the simulation is exact, and under Hamiltonian dynamics, the acceptance probability is 1). Thus, $p(\text{accept}|\epsilon)$ decreases with $\epsilon$.
Putting this idea together, we model $p(\text{accept}|\beta,\epsilon)$ as a logistic function for each $\beta\in\{0=\beta_1,...,\beta_J=1\}$
\begin{equation}
\text{logit}(p(\text{accept}|\beta,\epsilon))= w_0^{(j)}+w_1^{(j)}\epsilon
\end{equation}
Given data $\{(\beta^{(i)},y^{(i)})\}_{i=1,...,N}$ with $y^{(i)}=1$ if the proposed sample $i$ was accepted, and $y^{(i)}=0$ otherwise, we find
\begin{equation}
\label{eqn:propadapt}
\begin{aligned}
& \underset{\{w^{(j)}\}}{\text{max}}
& & \sum_{j=1}^J h(w^{(j)})\\
& \text{s.t.}
& & w^{(j)}_1\leq 0\\
& & & g(\beta_j,\epsilon) \leq g(\beta_{j-1},\epsilon)\,\,\, \forall\, \epsilon
\end{aligned}
\end{equation}
where
\begin{gather*}
h(w^{(j)})=\sum_{i:\beta^{(i)}=\beta_j} y^{(i)} \log(g(\beta^{(i)},\epsilon^{(i)}))\\+(1-y^{(i)})\log(1-g(\beta^{(i)},\epsilon^{(i)}))
\end{gather*}
and $$g(\beta_j,\epsilon)=p(\text{accept}|\beta_j,\epsilon)=\frac{1}{1+\exp(-(w^{(j)}_0+w^{(j)}_1\epsilon))}$$
The last constraint can be satisfied by enforcing $g(\beta_j,\epsilon_\text{min})\leq g(\beta_{j-1},\epsilon_\text{min})$ and $g(\beta_j,\epsilon_\text{max})\leq g(\beta_{j-1},\epsilon_\text{max})$, as doing so will ensure $g(\beta_j,\epsilon)\leq g(\beta_{j-1},\epsilon)$ for all $\epsilon\in[\epsilon_\text{min},\epsilon_\text{max}]$. Before solving (\ref{eqn:propadapt}), we first run chains at fixed $\beta=0$ and $\beta=1$, running a basic stochastic optimization method to adapt each stepsize until the acceptance rate is close to the target acceptance rate, which we take to be 0.651, which is suggested by~\cite{hmc_tuning}. We take these stepsizes to be $\epsilon_\text{max}$ and $\epsilon_\text{min}$, respectively. Once we have approximated $p(\text{accept}|\beta,\epsilon)$, choosing the appropriate proposal distribution given $\beta$ is simple: $$\hat{\epsilon}_\text{opt}(\beta_j)=\frac{\text{logit}(\text{p(acc)})-w^{(j)}_0}{w^{(j)}_1}$$
If $\hat{\epsilon}_\text{opt}$ is outside $[\epsilon_\text{min},\epsilon_\text{max}]$, we project it into the interval.

\subsection{Example}
\label{sec:gaussian}

Here we consider a target distribution of a mixture of two 10-dimensional Gaussians, each having a covariance of $0.5I$ separated in the first dimension by 5. Our prior distribution for the interpolating scheme is a zero mean Gaussian with covariance $30I$. The prior was chosen by looking at a one-dimensional projection of the target distribution and picking a zero-mean prior whose variance, $\sigma^2$, adequately covered both of the modes. The variance of the multidimensional prior was taken to be $\sigma^2 I$, and the mean to be $\textbf{0}$. Our prior on temperatures was taken to be uniform. We compare the adaptive method above to simulation with a fixed step size, which is determined by averaging all of the step sizes, in an effort to pick the optimal fixed step size. The below figures show an improvement over the fixed step size in mixing and partition function estimation using our adaptive scheme.

We obtained similar improvements using random walk Metropolis by varying the covariance of an isotropic Gaussian proposal distribution. We note another scheme for discrete binary data may be used, where the number of variables in the target distribution to ``flip", as a function of temperature, is a parameter.

%% file: sub/uniformExperiment.tex
\begin{figure*}[ht]
\includegraphics[width=.33\textwidth]{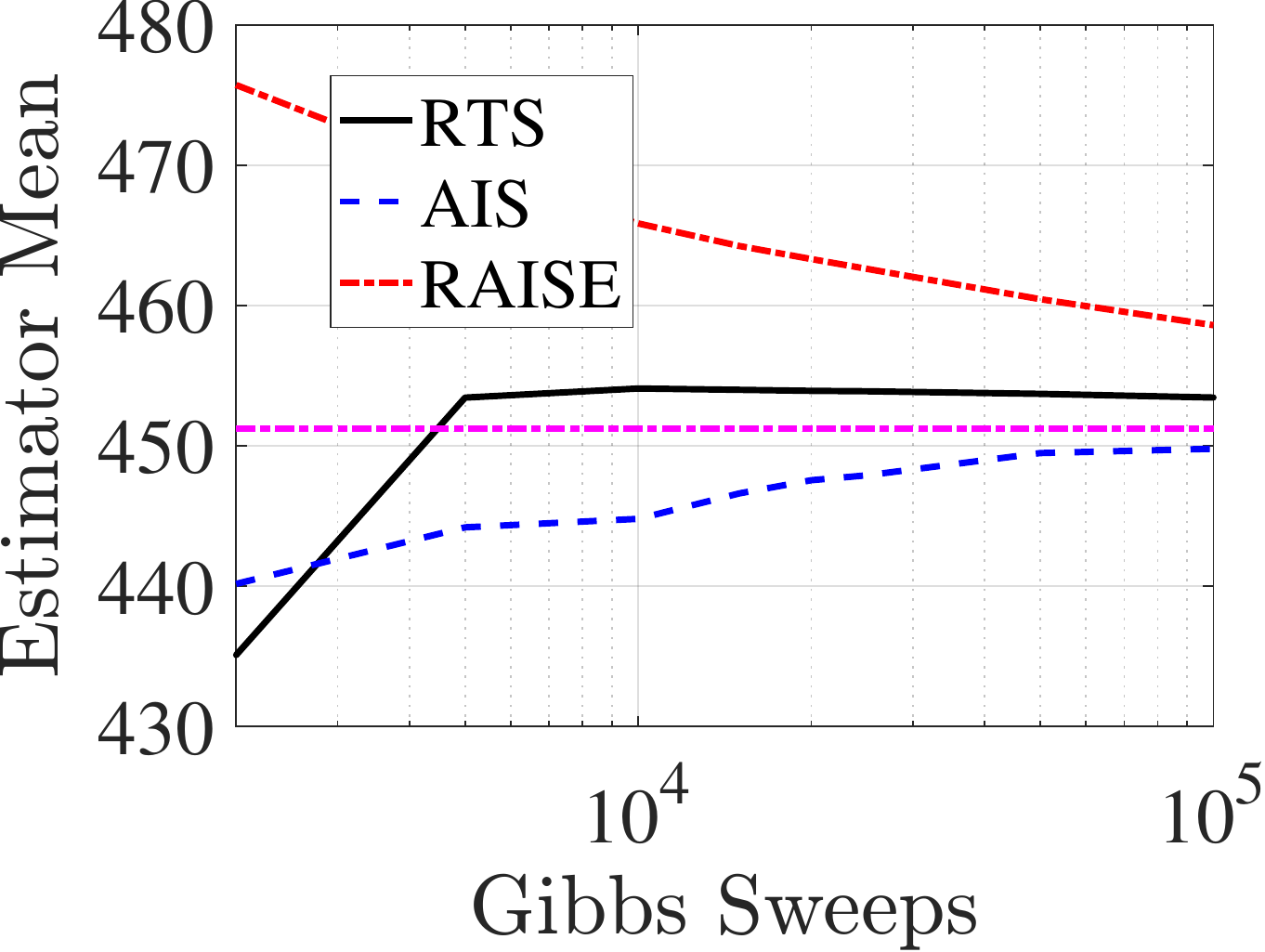}
\includegraphics[width=.33\textwidth]{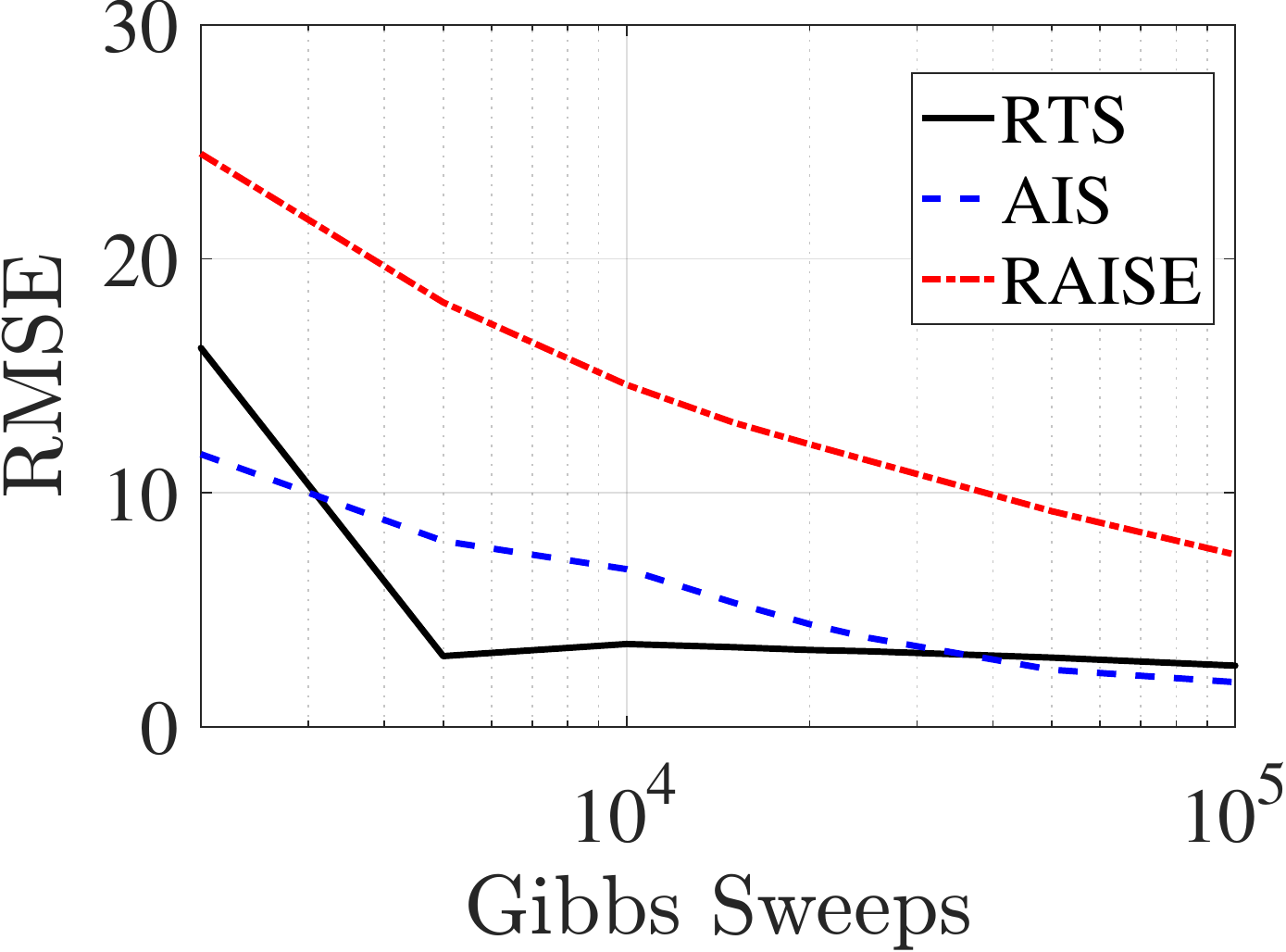}
\includegraphics[width=.33\textwidth]{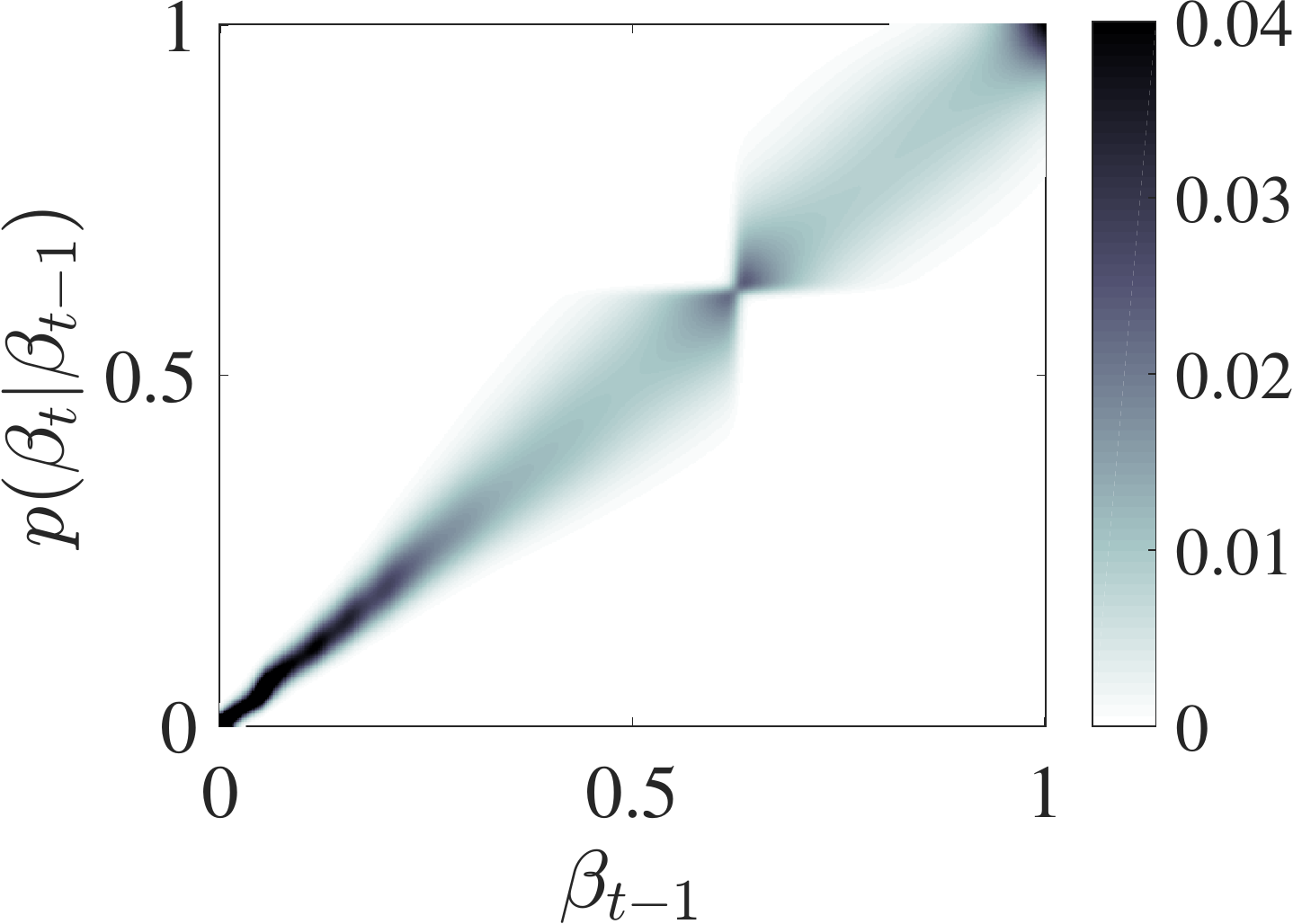}
\caption{\label{fig:uniform} $\log Z$ estimates for an RBM with 784 visible units and 500 hidden units trained on the MNIST dataset when $p_1$ is a uniform distribution.  (Left) The mean of the competing estimators.  The magenta line gives truth.  (Middle) The RMSE of the competing estimators.  (Right) The empirical transition matrix on $\beta$ clearly demonstrates that there is a ``knot'' in the temperature distribution that is prohibiting effective mixing and reducing estimator quality.  This gives a simple diagnostic to analyze sampling results and mixing properties.}
\end{figure*}
\section{RBM $\log Z$ Estimates from a Uniform $p_1$}
\label{sec:uniform}
The choice of $p_1$ is known to dramatically affect the quality of log partition function estimates, and this was noted for RBMs in \cite{salakhutdinov2008quantitative}.  To demonstrate the comparative effect of a poor $p_1$ distribution on our estimator, we choose $p_1$ to have a uniform distribution over all binary patterns, and follow the same experimental setup as in Section \ref{sec:pf_rbm}.  The quantitative results are shown in Figure \ref{fig:uniform} (Left) and (Middle).  In this case all estimators behave significantly worse than when $p_1$ was intelligently chosen.  We note that the initialization stage of RTS (see Section \ref{sec:burnin}) takes significantly longer with this choice of $p_1$.  Initially RTS decreases bias faster than AIS, but asymptotically they have similar behavior up to $10^5$ Gibbs sweeps.  

The poor performance of the estimators is due to a ``knot'' in the interpolating distribution caused by the mismatch between $p_1$ and $p_K$.  This can be clearly seen in the empirical transition matrix over the inverse temperature $\beta$, shown in Figure \ref{fig:uniform} (Right).  While we have limited our experiments to the interpolating distribution, a strength of our approach is that can naturally incorporate other possibilities that ameliorate these issues, such as moment averaging~\cite{grosse2013annealing} or tempering by subsampling \cite{van2014tempering}, as mentioned in Section \ref{sec:st}.


%% file: sub/transitionMatrix.tex
\begin{figure*}
\centering
\includegraphics[width=.45\textwidth]{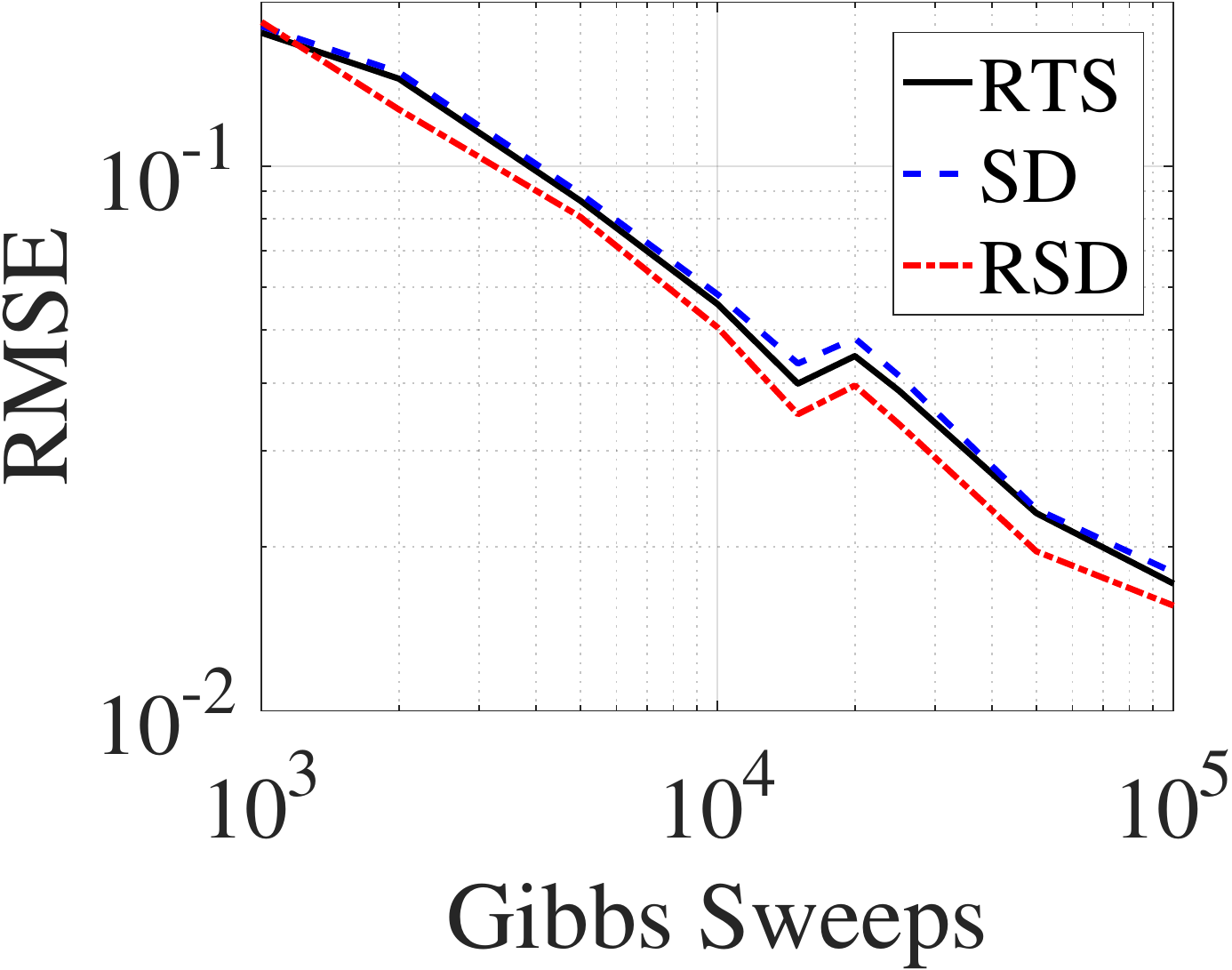}
\includegraphics[width=.45\textwidth]{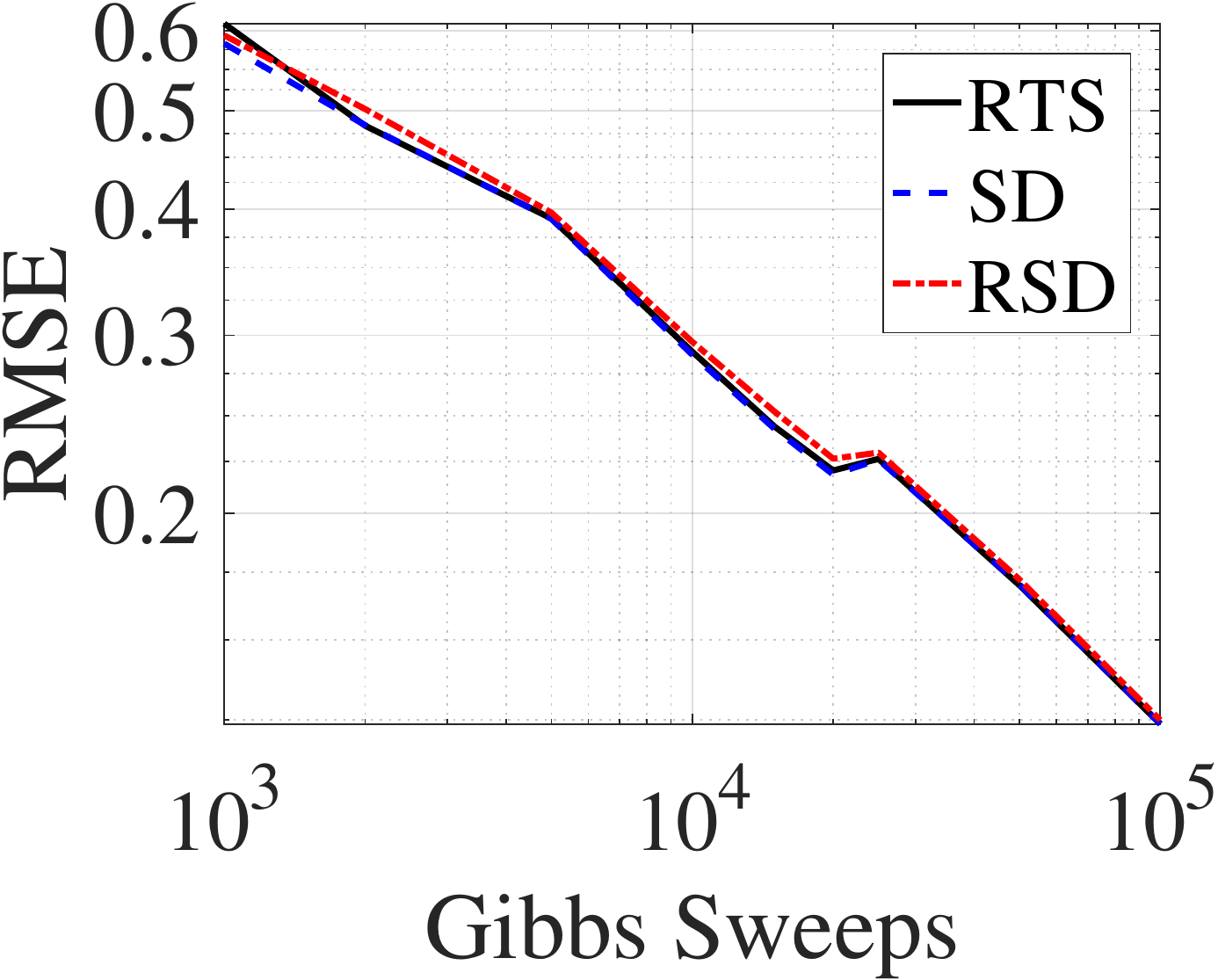}
\caption{\label{fig:rtm} An illustration of the effect of estimating the stationary distribution from the transition matrix.  Both plots show the RMSE on RBMs averaged over 20 repeats.  Experimental procedure is the same as the main text. (Left)  RTS, TM, and RTM compared on a 784-10 RBM.  Because the latent dimensionality is small, mixing is very effective and accounting for the transition matrix improves performance consistently by about 10\%. (Right) For an 784-200 RBM, the approximation as a Markov transition is inaccurate, and we observe no performance improvements.}
\end{figure*}

\section{Estimating $q(\beta_k)$ from a transition matrix}
\label{sec:tm}
Instead of estimating $q(\beta_k)$ by Rao-Blackwellizing via $c_k$ in \eqref{ck}, it is possible to estimate $q(\beta_k)$ from the stationary distribution of a transition matrix.  
The key idea here is that the transition matrix accounts for the sampling structure used in MCMC algorithms, whereas $c_k$ is derived using i.i.d. samples.
Suppose that we have a transition sequence $\beta_1 \rightarrow \beta_2 \dots \rightarrow \beta_N$.
If $p(x|\beta)$ is an exact Gibbs sampler, then this is a Markov transition, since
\begin{align}
p(\beta_{n+1}&=\beta_k| \beta_{n}=\beta_j),\nonumber\\
&=\sum_{x} p(\beta_{n+1}=\beta_k|x)p(x|\beta_{n}=\beta_j),\nonumber\\
&=P_{jk}.\nonumber
\end{align}
Note that in general that we do not have an exact Gibbs sampler on $p(x|\beta)$.  In these cases the approach is approximate.  The top eigenvector of $P$ gives the stationary distribution over $\beta_k$, which is $q(\beta_k)$.  We briefly mention two importance sampling strategies to estimate this transition matrix.  First, this matrix can simply be estimated with empirical samples, with
$$ P_{jk}\propto \sum 1_{\{\beta_n+1=\beta_k, \beta_n=\beta_j\}},$$
where $1_{\{\cdot\}}$ is the identity function.  Then $q(\beta_k)$ is estimated from the top eigenvector.  We denote this strategy Stationary Distribution (SD).  
A second approach is to Rao-Blackwellize over the samples, where
$$ P_{jk}\propto \sum p(\beta_n+1=\beta_k|x_n)1_{\{\beta_n=\beta_j\}} .$$  We denote this strategy as Rao-Blackwellized Stationary Distribution (RSD).

The major drawback of this approach is that it is rare to have exact Gibbs samples over $p(x|\beta)$, but instead we have a transition operation $T(x_n|\beta,x_{n-1})$.  
In this case, it is unclear whether this approach is useful.  We note that in simple cases, such as a RBM with 10 hidden nodes, RSD can sizably reduce the RMSE over RTS, 
as shown in Figure \ref{fig:rtm}(Left).  However, in more complicated cases when the assumption that we have a Gibbs sampler over $p(x|\beta)$ breaks down, there is essentially no change between RTS and RSD, as shown in a 200 hidden node RBM in Figure \ref{fig:rtm} (Right).  
Our efforts to correct the transition matrix for the transition operator instead of a Gibbs sampler did not yield performance improvements.